\documentclass[final]{cvpr}
\usepackage{times}
\usepackage[ruled,vlined]{algorithm2e}
\usepackage[export]{adjustbox}
\usepackage{amsmath}
\usepackage{amssymb}
\usepackage{array,multirow}
\usepackage{booktabs}
\usepackage{caption}
\usepackage{subcaption}
\usepackage{cuted}
\usepackage{graphicx}
\usepackage{overpic}
\usepackage{xcolor}
\usepackage[pagebackref=true,breaklinks=true,letterpaper=true,colorlinks,bookmarks=false]{hyperref}
\usepackage{cleveref}

\def\cvprPaperID{552}
\def\confYear{CVPR 2021}

\newcommand{\method}[0]{NeuroMorph\xspace}

\title{\method: Unsupervised Shape Interpolation\\ and Correspondence in One Go}
\author{Marvin Eisenberger$^{*,\dag}$, David Novotny$^{*}$, Gael Kerchenbaum$^{*}$,  Patrick Labatut$^{*}$, \\   Natalia Neverova$^{*}$, Daniel Cremers$^{\dag}$, Andrea Vedaldi$^{*}$\\[5pt]
Facebook AI Research$^{*}$, Technical University of Munich$^{\dag}$
}

\makeatletter
\renewcommand{\paragraph}{%
  \@startsection{paragraph}{4}%
  {\z@}{.5ex \@plus 1ex \@minus .2ex}{-1em}%
  {\normalfont\normalsize\bfseries}%
}
\makeatother

\begin{document}
\maketitle
\begin{strip}
\vspace*{-3em}\\
\includegraphics[width=\linewidth]{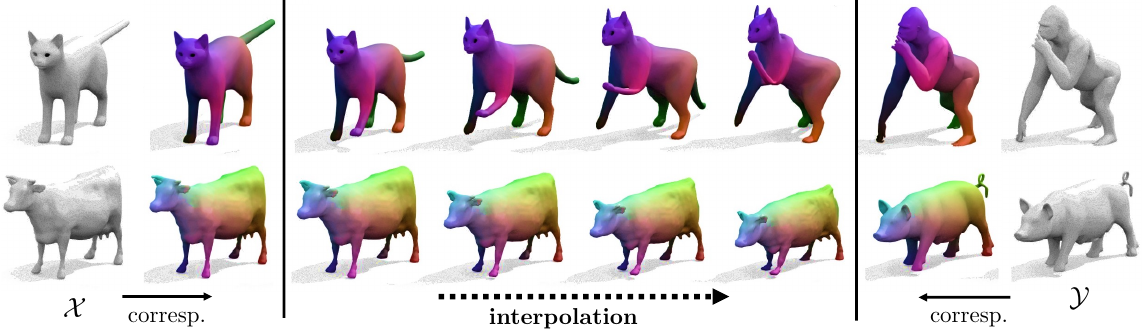}
\captionof{figure}{
Our \textbf{\method} neural network takes as input two meshes (left and right) and produces in one go (i.e.~in a single feed-forward pass) a continuous \textbf{interpolation} and point-to-point \textbf{correspondence} between them (color coded).
The interpolation, expressed as a displacement field, changes the pose of the source shape while preserving its identity.
}\label{f:spalsh}
\end{strip}

\thispagestyle{empty} \pagestyle{empty}

\begin{center}
  \large\textbf{Abstract}
\end{center}

We present \method, a new neural network architecture 
that takes as input two 3D shapes and produces in one go, i.e.~in a single feed forward pass, a smooth interpolation and point-to-point correspondences between them.
The interpolation, expressed as a deformation field, changes the pose of the source shape to resemble the target, but leaves the object identity unchanged.
\method uses an elegant architecture combining graph convolutions with global feature pooling to extract local features.
During training, the model is incentivized to create realistic deformations by approximating geodesics on the underlying shape space manifold.
This strong geometric prior allows to train our model end-to-end and in a fully unsupervised manner without requiring any manual correspondence annotations.
\method works well for a large variety of input shapes, including non-isometric pairs from different object categories.
It obtains state-of-the-art results for both shape correspondence and interpolation tasks, matching or surpassing the performance of recent unsupervised and supervised methods on multiple benchmarks.

\section{Introduction}\label{s:intro}


The ability to relate the 3D shapes of objects is of key importance to fully understand object categories.
Objects can change their shape due to articulation, other motions and intra-category variations, but such changes are not arbitrary.
Instead, they are strongly constrained by the category of the objects at hand.
Seminal works such as~\cite{loper15smpl:} express such constraints by learning statistical shape models.
In order to do so, they need to put in correspondence large collections of individual 3D scans, which they do by exploiting the fact that individual objects deform continuously in time, and by using some manual inputs to align different object instances.
%
%
Due to the high complexity of obtaining and pre-processing such 3D data, however, these models remain rare and mostly limited to selected categories such as humans that are of sufficient importance in applications.
In this paper, we are thus interested in developing a method that can learn to relate different 3D shapes fully automatically, interpolating a small number of 3D reconstructions, and in a manner which is less specific to a single category (\Cref{f:spalsh}).

Due to the complexity of this task, authors have often considered certain sub-problems in isolation.
One is to establish \emph{point-to-point correspondences} between shapes, telling which points are either physically identical (for a given articulated object) or at least analogous (for similar objects).
A second important sub-problem is \emph{interpolation}, which amounts to continuously deforming a source shape into a target shape.
Interpolation must produce a collection of intermediate shapes that are meaningful in their own right, in the sense of being plausible samples from the underlying shape distribution.
The interpolation \emph{trajectory} must be also meaningful; for instance, if the deformation between two shapes can be explained by the articulation of an underlying physical object, this solution is preferred.

The correspondence and interpolation problems have been addressed before extensively, by using tools from geometry and, more recently, machine learning.
Most of the existing algorithms, however, require at least some manual supervision, for example in the form of an initial set of sparse shape correspondences.
Furthermore, correspondence and interpolation are rarely addressed together due to their complexity.

In this paper, we advocate instead for an approach in which the correspondence and interpolation problems are solved \emph{simultaneously}, and in an \emph{unsupervised} manner.
To do this, we introduce \emph{\method}, a new neural network that solves the two problems in a single feed forward pass.
We show that, rather than making learning more difficult, integrating two goals reinforces them, making it possible to obtain excellent empirical results.
Most importantly, we show that \method can be learned in a \emph{fully unsupervised} manner, given only a collection of 3D shapes as input and certain geometric priors for regularization.

\method advances the state of the art in shape matching and interpolation, surpassing by a large margin prior unsupervised methods and often matching the quality of supervised ones.
We show that \method can establish high-quality point-to-point correspondences without any manual supervision even for difficult cases in which shapes are related by substantial non-isometric deformations (such as between two different types of animals, like a cat and a gorilla, as in~\Cref{f:spalsh}) which have challenged prior approaches.
Furthermore, we also show that \method can interpolate effectively between different shapes, acting on the \emph{pose} of a shape while leaving its \emph{identity} largely unchanged.
To demonstrate the quality of the interpolation, we use it for \emph{data augmentation}, extending a given dataset of 3D shapes with intermediate ones.
Augmenting a dataset in this manner is useful when, as it is often the case, 3D training data is scarce.
We show the benefits of this form of data augmentation to supervise other tasks, such as reconstructing continuos surfaces from sparse point clouds.

Our new formulation also gives rise to some interesting applications:
Since our method learns a function that produces correspondence and interpolation in a single feed forward pass, it can be used not only to align different shapes, but also for pose transfer, digital puppeteering and other visual effects.

\section{Related work}\label{s:related}

To the best of our knowledge, we are the first to consider the problem of learning a mapping that, given a pair of shapes as input, predicts in a feed-forward manner their correspondences and interpolation.
This should be contrasted to other recent approaches to shape understanding such as LIMP~\cite{cosmo2020limp} that try to learn a shape space.
These architectures need to solve the difficult problem of \emph{generating} or \emph{auto-encoding} 3D shapes.
Unfortunately, designing good generator networks for 3D shapes remains a challenging problem.
In particular, it is difficult for these networks to generalize beyond the particular family of shapes (e.g.~humans) experienced during training.
By contrast, we do not try to generate shapes outright, but only to \emph{relate} pairs of given input shapes.
This replaces the difficult task of shape generation with the easier task of generating a deformation field, working well with a large variety of different shapes.

The rest of the section discusses other relevant work.

\paragraph{Shape correspondences.}

The problem of establishing correspondences between 3D shapes has been studied extensively (see the recent surveys~\cite{vankaick11correspsurvey,tam2013pointcloudsurvey,sahilliouglu2019recent}).
Traditional approaches define axiomatic algorithms that focus on a certain subclass of problems like rigid transformations~\cite{yang2015go,zhou2016fastglobal}, nearly-isometric deformations~\cite{ovsjanikov2012functional,aflalo2016spectral,vestner2017pmf,ren2018orientation}, bounded distortion~\cite{melzi2019zoomout,eisenberger2020smooth} or partiality~\cite{litany2016puzzles,rodola2016partial,litany2017fullyspectral}.
Methods such as functional maps~\cite{ovsjanikov2012functional} reduce matching to a spectral analysis of 3D shapes.

More recent approaches use machine learning and are often based on developing deep neural networks for non-image data such as point clouds, graphs and geometric surfaces~\cite{bronstein2017geometric}.
Charting-based methods define learnable intrinsic patch operators for local feature aggregation~\cite{masci2015geodesic,boscaini2016learning,monti2017geometric,poulenard2018multi,Wiersma2020}.
Deep functional maps~\cite{litany2017deep} aim at combining a learnable local feature extractor with a differentiable matching layer based on the axiomatic functional maps framework~\cite{ovsjanikov2012functional}.
Subsequent works~\cite{halimi2019unsupervised,roufosse2019unsupervised} extended this idea to the unsupersived setting and combined it with learnable point cloud feature extractors~\cite{donati2020deep,sharma2020weakly}.
%
%
Moreover,~\cite{eisenberger2020deep} recently proposed to replace the functional maps layer with a multi-scale correspondence refinement layer based on optimal transport.
Another related approach is~\cite{groueix20183dcoded} which uses a PointNet~\cite{qi2017pointnet} encoder to align a human template to point cloud observations to compute correspondences between different human shapes.

\paragraph{Feature extractors for 3D shapes.}


Several authors have proposed to reduce matching 3D shapes to matching local shape descriptors.
A common remedy is learning to refine hand-crafted descriptors such as SHOT~\cite{tombari2010SHOT}, e.g. with metric learning~\cite{litany2017deep,halimi2019unsupervised,roufosse2019unsupervised}. 
In practice, this approach is highly dependent on the quality of the input features and tends to be unstable due to the noise and the complex variable structures of real 3D data.
%
More recently, authors have thus looked at \emph{learning} such descriptors directly~\cite{donati2020deep,sharma2020weakly} with point cloud feature extractors~\cite{thomas2019kpconv,qi2017pointnet++}.
Another possibility is to interpret a 3D mesh as a graph and use graph convolutional neural networks~\cite{kipf2016semi,defferrard2016convolutional}.
The challenge here is that the specific graph used to represent a 3D shape is partially arbitrary (because we can triangulate a surface in many different ways), and graph convolutions must discount geometrically-irrelevant changes (this is often done empirically by re-meshing as a form of data augmentation).

\paragraph{Shape spaces, manifolds and interpolation.}

3D shapes can be interpreted as low-dimensional manifolds in a high-dimensional embedding space~\cite{kilian2007geometric,wirth2011shapespace,heeren2012time,heeren2016shellsplines}.
The low-dimensional manifold can, for example, capture the admissible pose of an articulated object~\cite{zhang2015shell,von2016optimized,heeren2018principal}.
Given a shape manifold, interpolation can then be elegantly formulated as finding geodesic paths between two shapes.
However, building shape manifolds may be difficult in practice, especially if the input shapes are not in perfect correspondence.
Therefore, also inspired by LIMP~\cite{cosmo2020limp}, for training \method we follow approaches such as~\cite{eisenberger2019divergence,eisenberger2020hamiltonian} that avoid building a shape manifold explicitly and instead directly construct geodesic paths that originate at the source shapes and terminate in the vicinity of the target shapes.


\paragraph{Generative shape models.}

While manifolds provide a geometric characterization of a shape space, generative models provide a statistical one.
One particular challenge in this context is designing shape-decoder architectures that can generate 3D surfaces from a latent shape representation. A straightforward solution is predicting occupancy probabilities on a 3D voxel grid~\cite{choy20163d}, but the cost of dense, volumetric representations limits the resolution. Other approaches decode point clouds~\cite{fan2017point,yang2018foldingnet} or 3D meshes~\cite{groueix2018papier,gkioxari2019mesh} directly. 
A recent trend is encoding an implicit representation of a 3D surface in a neural network~\cite{mescheder2019occupancy,park2019deepsdf}. This allows for a compact shape representation and a decoder that can generate shapes of an arbitrary topology.
Following the same methodology, \cite{niemeyer2019occupancy} predicts a time-dependent displacement field that can be used to interpolate 3D shapes. This approach is related to ours, but it requires 4D supervision during training, whereas our method is trained on a sparse set of poses. 
ShapeFlow~\cite{jiang2020shapeflow} predicts dense velocity fields for template-based reconstruction. 
Similarly,~\cite{wang20193dn} computes an intrinsic displacement field to align a pair of input shapes, but they do not predict an intermediate sequence.

\section{Method}\label{s:method}

\newcommand{\p}{\x}
\newcommand{\q}{\y}
\newcommand{\x}{\mathbf{x}}
\newcommand{\y}{\mathbf{y}}
\newcommand{\X}{\mathbf{X}}
\newcommand{\Y}{\mathbf{Y}}
\newcommand{\fX}{\tilde{\X}}
\newcommand{\fY}{\tilde{\Y}}
\newcommand{\bv}{\mathbf{v}}
\newcommand{\fx}{\tilde{\x}}
\newcommand{\fy}{\tilde{\y}}
\newcommand{\rx}{\tilde{\x}'}
\newcommand{\ry}{\tilde{\y}'}
\newcommand{\bV}{\mathbf{V}}
\newcommand{\bZ}{\mathbf{Z}}

Let $\mathcal{X}$ and $\mathcal{Y}$ be 3D shapes, respectively called the \emph{source} and the \emph{target}, expressed as triangular meshes with vertices $\X=(\p_i)_{1\leq i\leq n}\in\mathbb{R}^{n\times 3}$ and $\Y = (\q_j)_{1\leq j\leq m}\in \mathbb{R}^{m\times 3}$, respectively.
Our goal is to learn a function
$$
f ~:~ (\mathcal{X},\mathcal{Y}) \longmapsto (\Pi, \Delta),
$$
that, given the two shapes as input, predicts `in one go' a correspondence matrix $\Pi$ and an interpolation flow $\Delta$ between them.
The matrix $\Pi \in [0,1]^{n\times m}$ sends probabilistically the vertices $\p_i$ of the source mesh $\mathcal{X}$ to corresponding vertices $\q_j$ in the target mesh $\mathcal{Y}$ and is thus row-stochastic (i.e.~$\Pi \mathbf{1} = \mathbf{1}$).
The interpolating flow $\Delta(t) \in \mathbb{R}^{n\times 3}$, $t\in [0,1]$ shifts continuously the vertices of the source mesh, forming trajectories:
\begin{equation}\label{eq:displacement}
	\X(t):=\X+\Delta(t),
\end{equation}
that take them from their original locations $\X(0) = \X$ to new locations $\X(1) \approx \Pi \Y$ close to the corresponding vertices in the target mesh.

The function $f$ is given by two deep neural networks.
The first, discussed in~\Cref{subsec:feat}, establishes the correspondence matrix $\Pi$ and the second, discussed in~\Cref{subsec:displacement}, outputs the shifts $\Delta(t)$ for arbitrary values of $t\in[0,1]$.
Both networks are trained end-to-end in an unsupervised manner, as described in~\Cref{subsec:losses}.

\subsection{Correspondences and vertex features}\label{subsec:feat}

The correspondence matrix $\Pi$ between meshes $\mathcal{X}$ and $\mathcal{Y}$ is obtained by extracting and then matching features of the mesh vertices.
The features are computed by a deep neural network:
$
	\fX = \Phi(\mathcal{X}) \in \mathbb{R}^{n\times d}
$
that takes the shape $\mathcal{X}$ as input and outputs a matrix $\fX=(\fx_i)_{1\leq i \leq n}$ with a feature vector $\fx_i$ for each vertex $i$ of the mesh.
Given analogous features $\fY =(\fy_i)_{1\leq i \leq m}\in\mathbb{R}^{m\times d}$ for the target shape, the correspondence matrix is obtained by comparing features via the cosine similarity and normalizing the rows using the softmax operator:
\begin{equation}\label{eq:match}
	\Pi_{ij}
	:=
	\frac
	{\exp(\sigma s_{ij})}
	{\sum_{k=1}^{m}\exp(\sigma s_{ik})}
	\text{~~s.t.~~}
	s_{ij}
	:=
	\frac
	{\langle\fx_i,\fy_j\rangle_2}
	{\|\fx_i\|_2\|\fy_j\|_2},
\end{equation}
with temperature $\sigma \in \mathbb{R}^+$. In this way, $\Pi\mathbf{1}=\mathbf{1}$ and $\Pi$ can be interpreted as a soft assignment of source vertices $\p_i$ in $\mathcal{X}$ to target vertices $\q_j$ in $\mathcal{Y}$.

\paragraph{Feature extractor network.}

\begin{figure}
\includegraphics[width=\linewidth]{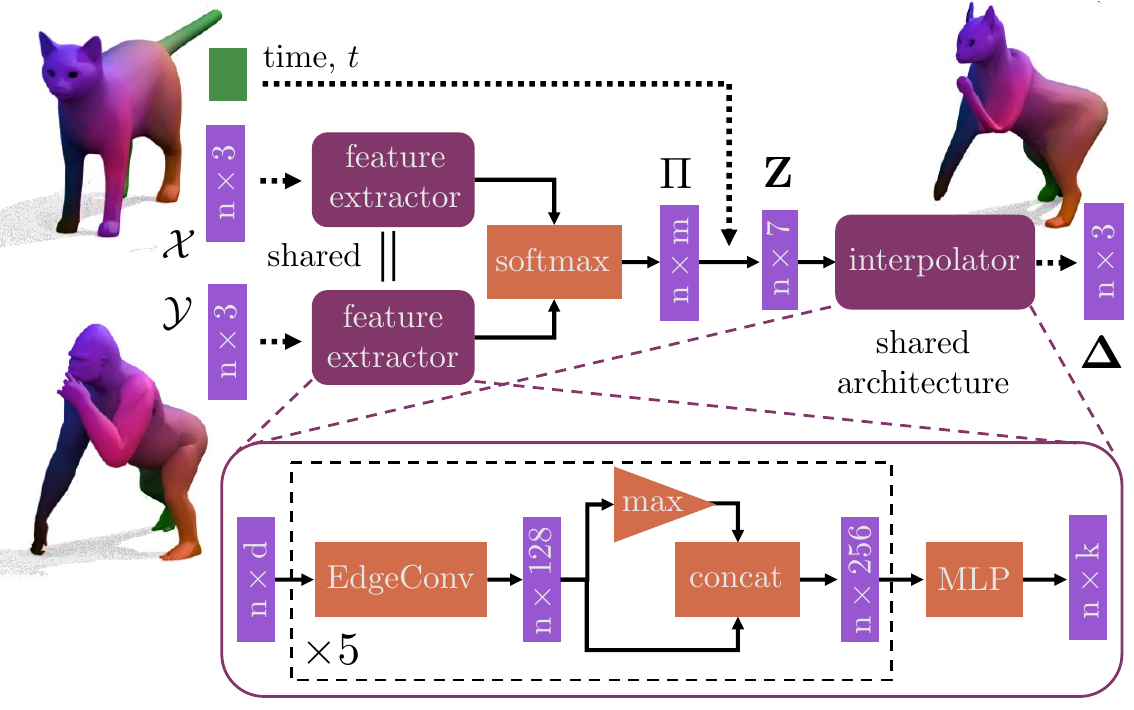}
\caption{\textbf{\method.} An overview of our model.}\label{fig:eclayer}
\end{figure}

Next, we describe the neural network $\Phi(\mathcal{X})$ that extracts the feature vectors that appear in~\Cref{eq:match} (this network is also illustrated in~\Cref{fig:eclayer}).
While different designs are conceivable, we propose here one based on successive local feature aggregation and global feature pooling.
The architecture makes use of the mesh vertices $\X$ as well as the mesh topology.
The latter is specified by the neighborhood structure $\mathcal{E}$, where $(i,j)\in\mathcal{E} \subset \{1,\dots,n\}^2$ means that vertex $\p_j$ is connected to vertex $\p_i$ by a triangle edge.
Thus, the mesh is fully specified by the pair $\mathcal{X}=(\X,\,\mathcal{E})$.

The layers of the network $\Phi$ are given by EdgeConv~\cite{wang2019dynamic} graph convolution operators implemented via residual sub-networks~\cite{he15deep}.
In more detail, each EdgeConv layer takes as input vertex features $\fX = (\fx_i)_{1\leq i\leq n}$ on $\mathcal{X}$ and computes an improved set of features $\fX' = (\fx'_i)_{1\leq i\leq n}$ via the expression:
\begin{equation}\label{eq:edgeconv}
	\fx'_i:=\max_{j:(i,j)\in\mathcal{E}}h_\phi(\fx_i,\fx_j-\fx_i).
\end{equation}
Here, a small residual network $h_\phi$ is used to combine the feature $\fx_i$ of the $i$-th vertex with the feature $\fx_j-\fx_i$ of one of the edges adjacent to it.
This is repeated for all edges incident on the $i$-th vertex and the results are aggregated via component-wise max-pooling over the mesh neighborhood $\bigl\{j:(i,j)\in\mathcal{E}\bigr\}$, resulting in an updated vertex feature $\fx_i'$.

The EdgeConv layer can effectively learn the local geometric structures in the vicinity of a point.
However, that alone is not sufficient to resolve dependencies in terms of the global geometry, since the message passing only allows for a local information flow.
Therefore, we append a global feature vector to the point features after each EdgeConv refinement, by applying the max pooling operator globally:
\begin{equation}\label{eq:appendglobal}
	\fx_i'' := \bigl(\fx_i',\max_{1\leq i\leq n}\fx_i'\bigr).
\end{equation}

The network $\Phi$ is given by a succession of these layers, forming a 
chain $\fX \rightarrow \fX' \rightarrow \fX'' \rightarrow \dots$ alternating global~\eqref{eq:edgeconv} and local~\eqref{eq:appendglobal} update steps.
The input features $\fX = (\X, \mathbf{N})$ are given by the concatenation of the absolute position of the mesh vertices $\X$ with the outer normals at the vertices $\mathbf{N} = (\mathbf{n}_i)_{1\leq i \leq n}$ (the normal vectors $\mathbf{n}_i$ are computed by averaging over face normals adjacent to $\p_i$).

\subsection{Interpolator}\label{subsec:displacement}

We are now ready to describe the interpolator component of our model.
Recall that the goal is to predict a displacement operator $\Delta$ such that the trajectory $\X(t) = \X + \Delta(t)$ smoothly shifts the point of the first mesh to points in the second.
Notice that $\Delta(t) \in \mathbb{R}^{n\times 3}$ is just a collection of 3D vectors associated to each mesh vertex, just like the vertex positions, normals and feature vectors in the previous section.
Thus, we offload the calculation of the displacements to a similar convolutional neural network and write:
$
   \Delta(t) = \Psi(\mathcal{X},\mathcal{Y},t).
$
The difference is the input to the network $\Psi$, which is now given by the 7-dimensional feature vectors $\bZ\in\mathbb{R}^{n\times 7}$:
\begin{equation}
	\bZ:=\bigl(\X,\,\Pi \Y-\X,\,\mathbf{1} t\bigr).
\end{equation}
These feature vectors consist of the vertices $\X$ of the source shape $\mathcal{X}$, the offset vectors $\Pi \Y-\X$ predicted by the correspondence module of~\Cref{subsec:feat}\footnote{Note that $\Delta(1)=\Pi \Y-\X$.}, and the time variable $t$ (`broadcast' to all vertices by multiplication with a vector of all ones).
Just like the network $\Phi$ in~\Cref{subsec:feat}, the network $\Psi$ alternates global~\eqref{eq:edgeconv} and local~\eqref{eq:appendglobal} update steps to compute a sequence of updated features $\bZ \rightarrow \bZ' \rightarrow \bZ'' \dots \rightarrow \bV$ terminating in a matrix $\bV \in \mathbb{R}^{n\times 3}$.
The final displacements are then given by a scaled version of $\bV$, and are set to $\Delta(t) = t \bV(t)$.

In this manner, the network can immediately obtain a trivial (degenerate) solution to the interpolation problem by setting $\bV(t) = \Pi \Y - \X$, which amounts to copying verbatim part of the input features $\bZ$.
This result is a simple linear interpolation of the mesh vertices, trivially satisfying the boundary conditions of the interpolation:
\begin{align}
	\X(0) &= \X + 0 \cdot \bV(0) = \X,\label{e:begin}\\
	\X(1) &= \X + 1 \cdot (\Pi \Y - \X) = \Pi \Y\label{e:end}.
\end{align}
Linear interpolation provides a sensible initialization, but is in itself a degenerate solution as we wish to obtain `geometrically plausible' deformations of the mesh.
To prevent the network from defaulting to this case, we thus need to incentivize geometrically meaningful deformations during training, which we do in the next section.

\subsection{Learning}\label{subsec:losses}

In this section, we show how we can train the model in an \emph{unsupervised}%
\footnote{
	In practice, the only assumption we make about the input objects is that they are in an approximately canonical rigid pose in terms of the up-down and front-back orientation.
	For most existing benchmarks this holds trivially without any further preprocessing.
	The recent paper by~\cite{sharma2020weakly} calls this setup weakly supervised.}
\emph{manner}.
That is, given only a collection of example meshes with no manual annotations, our method simultaneously learns to interpolate and establish point-to-point correspondences between them.
This sets it apart from prior work on shape interpolation which either require dense correspondences during training or, in the case of classical axiomatic interpolation methods, even at test time.
Learning comprises three signals, encoded by three corresponding losses:
\begin{equation}\label{eq:loss}
	\ell:=
	\lambda_\mathrm{reg}\ell_\mathrm{reg}
	+\lambda_\mathrm{arap}\ell_\mathrm{arap}
	+\lambda_\mathrm{geo}\ell_\mathrm{geo}.
\end{equation}
The loss $\ell_\mathrm{reg}$ ensures that correspondences and interpolation correctly map the source mesh on the target mesh, and the other two ensure that this is done in a geometrically meaningful way.
The latter is done by constraining the trajectory $(\X(t))_{t\in[0,1]}$ generated by the model.
Recall that the model can be queried for an arbitrary value $t\in[0,1]$, and it is thus able to produce interpolations that are truly continuous in time.
During training, in order to compute our losses, we sample predictions $\X_0,\dots,\X_T$ for an equidistant set of discrete time steps
$
	\X_k:=\X(k/T)
$
where
$
		k=0,\dots,T.
$

\paragraph{Registration loss.}

Requirement~\Cref{e:begin} holds trivially as $\Delta(0)=0$ is built into our model definition (see \Cref{eq:displacement}).
For~\Cref{e:end}, we introduce the registration loss:
$
	\ell_\mathrm{reg}\bigl(\X_T,\Y,\Pi\bigr):=\|\Pi \Y-\X_T\|^2_2.
$
Since our goal is to compute shape interpolations without any supervision, we use the soft correspondences $\Pi$ estimated by our model instead of ground-truth annotations.

\paragraph{As-rigid-as-possible loss.}

In general, there are infinitely many conceivable paths between a pair of shapes.
In order to restrict our method to plausible sequences, we
regularize the path using the theory of shape spaces~\cite{kilian2007geometric,wirth2011shapespace,heeren2012time}.
%
As we work with discrete time, we approximate the `distance' between shapes in the shape space manifold by means of the local distortion metric between two consecutive states $\X_k$ and $\X_{k+1}$.
To that end, we choose the as-rigid-as-possible~\cite{sorkine2007rigid} metric:
\begin{multline*}
	E_\mathrm{arap}\bigl(\X_{k},\X_{k+1}\bigr):=\\
	\frac{1}{2}\min_{\substack{\mathbf{R}_i\in SO(3)\\i=1,\dots,n}}
	\sum_{(i,j)\in\mathcal{E}}
	\bigl\|\mathbf{R}_i(\X_{k,j}-\X_{k,i})-(\X_{k+1,j}
	-\X_{k+1,i})\bigr\|_2^2.
\end{multline*}

Intuitively, this functional rotates the local coordinate frame of each point in $\X_{k}$ to the corresponding deformed state $\X_{k+1}$ and penalizes deviations from locally rigid transformations.
Moreover, the rotation matrices $\mathbf{R}_i$ can be computed in closed form which allows for an efficient optimization of $E_\mathrm{arap}$ (see~\cite{sorkine2007rigid} for more details).
Finally, we can use this functional to construct the first component of our loss function for the whole sequence $(\X_k)_k$:
\begin{multline}
	\ell_\mathrm{arap}\bigl(\X_0,\dots,\X_T\bigr):=\\
	\sum_{k=0}^{T-1}E_\mathrm{arap}(\X_k, \X_{k+1})+E_\mathrm{arap}(\X_{k+1},\X_k).
\end{multline}


\paragraph{Geodesic distance preservation loss.}

The final component of our loss function in~\Cref{eq:loss} aims at preserving the pairwise geodesic distance matrices $\mathbf{D}_\mathcal{X}$ and $\mathbf{D}_\mathcal{Y}$ under the estimated mapping $\Pi$, this is given by
$
	\ell_\mathrm{geo}\bigl(\Pi\bigr):=\|\Pi \mathbf{D}_\mathcal{Y} \Pi^\top - \mathbf{D}_\mathcal{X}\|_2^2.
$
Note that this energy only regularizes the estimation of the correspondences $\Pi$ as $\mathcal{X}$ and $\mathcal{Y}$ are the (fixed) source and target shapes.

Intuitively, this objective promotes correspondences $\Pi$ with bounded geodesic distortion.
Variants of this objective are commonly used in classical shape matching~\cite{bronstein2006generalized,wang2011discrete,vestner2017pmf} and
have been also successfully integrated in a learning pipeline~\cite{halimi2019unsupervised} in combination with functional maps~\cite{ovsjanikov2012functional}.

\subsection{Implementation details}

\begin{table}
\scalebox{0.95}{
\begin{tabular}{lllll}
\toprule
&                                                 & err.         & p.p. & w/o p.p. \\
\midrule
\parbox[t]{2mm}{\multirow{3}{*}{\rotatebox[origin = c]{90}{\textit{Axiom.}}}}
& BCICP~\cite{ren2018orientation}                 & 6.4          & ---  & --- \\
& ZoomOut~\cite{melzi2019zoomout}                 & 6.1          & ---  & --- \\
&Smooth Shells~\cite{eisenberger2020smooth}       & 2.5          & ---  & --- \\
\midrule
\parbox[t]{2mm}{\multirow{3}{*}{\rotatebox[origin = c]{90}{\textit{Sup.}}}}
& 3D-CODED~\cite{groueix20183dcoded}              & 2.5          & ---  & ---  \\
&FMNet~\cite{litany2017deep}                      & 5.9          & PMF  & 11 \\
&GeoFMNet~\cite{donati2020deep}                   & 1.9          & ZO   & 3.1 \\
\midrule
\parbox[t]{2mm}{\multirow{3}{*}{\rotatebox[origin = c]{90}{\textit{Unsup.~~~~~~}}}}
& SurFMNet~\cite{roufosse2019unsupervised}        & 7.4          & ICP  & 15 \\
&Unsup. FMNet~\cite{halimi2019unsupervised}       & 5.7          & PMF  & 10  \\
&Weakly sup. FMNet~\cite{sharma2020weakly}        & 1.9          & ZO   & 3.3 \\
&Deep shells~\cite{eisenberger2020deep}           & 1.7          & ---  & ---  \\
&\method (Ours)                                   & \textbf{1.5} & SL   & \textbf{2.3} \\
\bottomrule
\end{tabular}
}
\centering
\caption{\textbf{Unsupervised correspondences on FAUST~\cite{Bogo:CVPR:2014} remeshed.}
Mean geodesic error in $\%$ of the diameter on the test set.
For methods that use an axiomatic technique for refinement (PMF~\cite{vestner2017pmf}, ZO~\cite{melzi2019zoomout}, ICP~\cite{ovsjanikov2012functional} or SL~\cite{eisenberger2020smooth}), we also show the result without.}\label{table:matchingaccuracy}

\end{table}
\begin{figure}
\includegraphics[width=\linewidth]{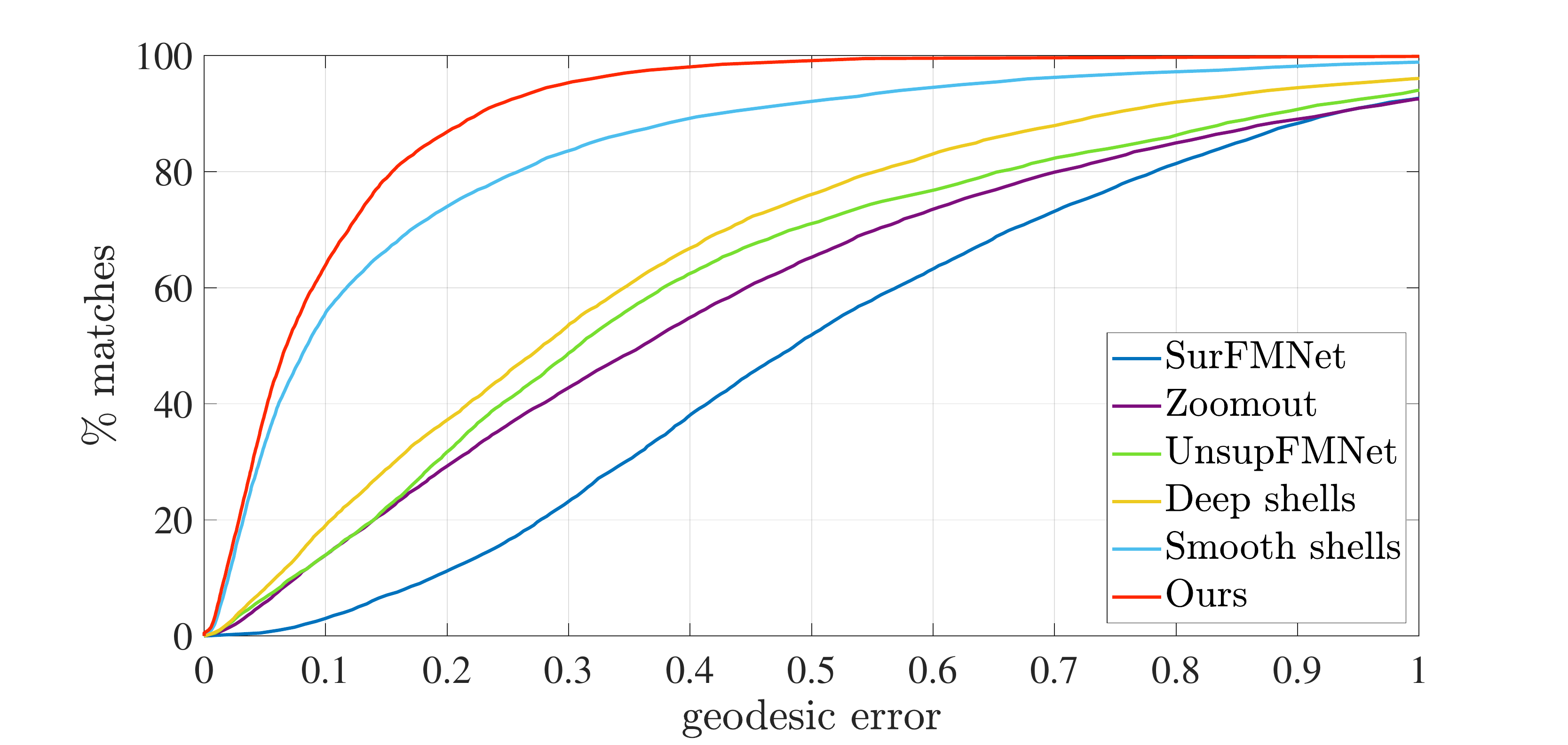}
\caption{\textbf{Unsupervised correspondences on SHREC20~\cite{dyke2020shrec}.}
We only compare our method to other unsupervised methods here, since there are no dense ground-truth correspondences for this benchmark which is a requirement for most supervised approaches.}\label{fig:shrec20comparison}
\end{figure}
\begin{figure*}
\centering
\begin{subfigure}{.25\linewidth}
  \centering
	\includegraphics[width=1.0\linewidth]{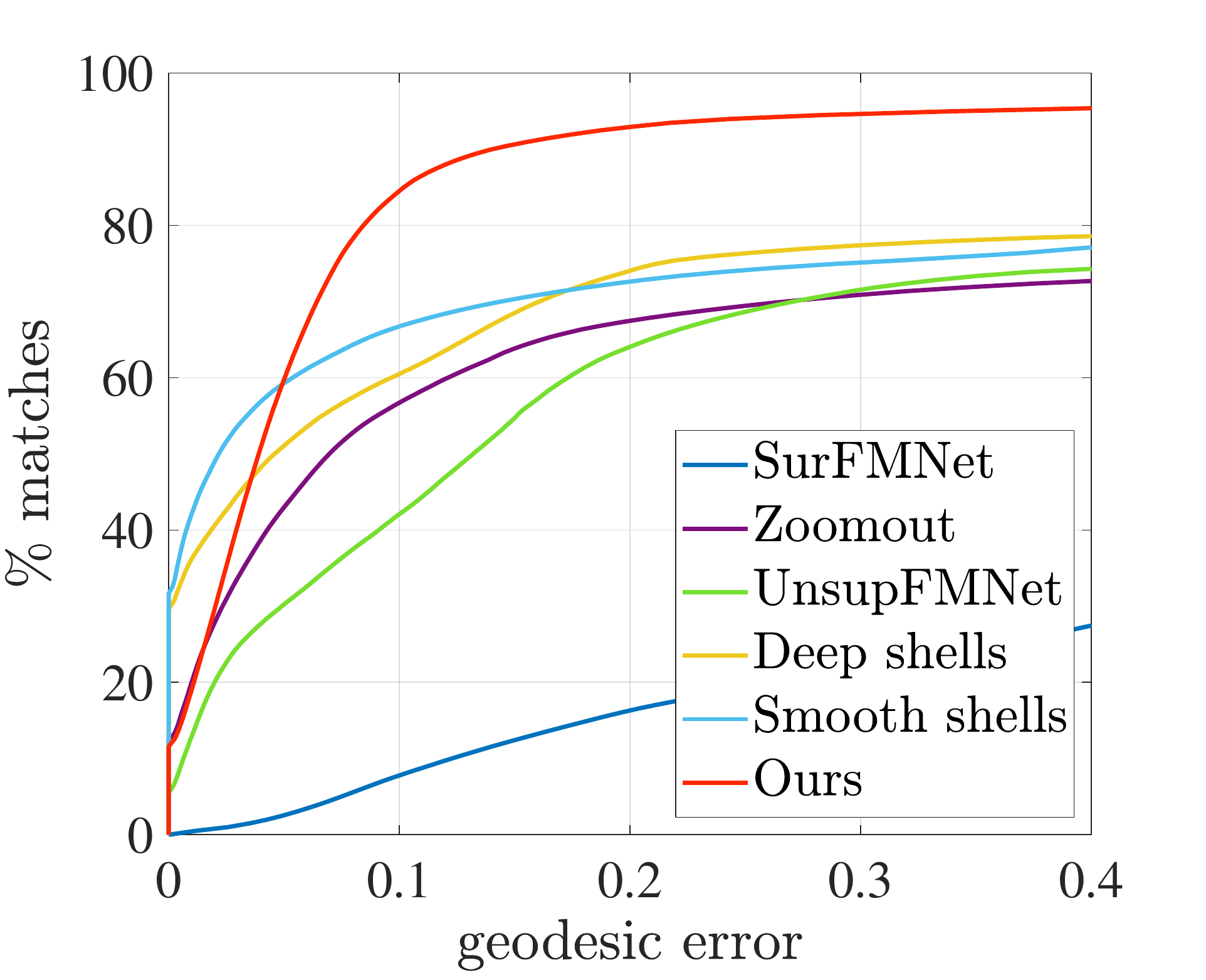}
\end{subfigure}
\begin{subfigure}{.74\linewidth}
\footnotesize
\begin{overpic}
        [width=0.99\linewidth]{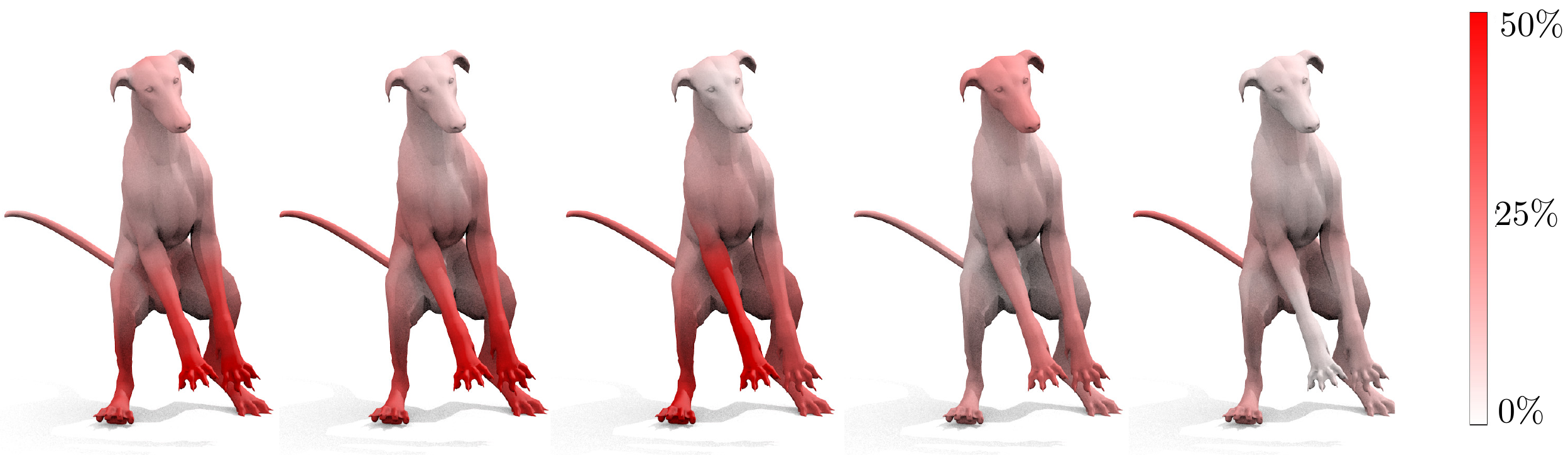}
        \put(3,28){ZoomOut~\cite{melzi2019zoomout}\small}
        \put(20,28){UnsupFMNet\cite{halimi2019unsupervised}\small}
        \put(39,28){Deep Shells~\cite{eisenberger2020deep}\small}
        \put(57,28){Smooth Shells~\cite{eisenberger2020smooth}\small}
        \put(80,28){Ours\small}
\end{overpic}
  \centering
\end{subfigure}
\caption{\textbf{Unsupervised correspondences on G-S-H.} We provide the cumulative geodesic error curves (in \% of the diameter) of different approaches (left). For a detailed comparison, we display heat maps on one pose of the 'Galgo' shape from our dataset (right). We color code the mean geodesic error for each point of the surface, averaged over all 1024 pairs from the test set. Our method is particularly good at discovering structural correspondences, i.e. matching extremities correctly.}\label{fig:heatmaps}
\end{figure*}
\begin{figure*}
\centering
\begin{overpic}
        [width=0.82\linewidth]{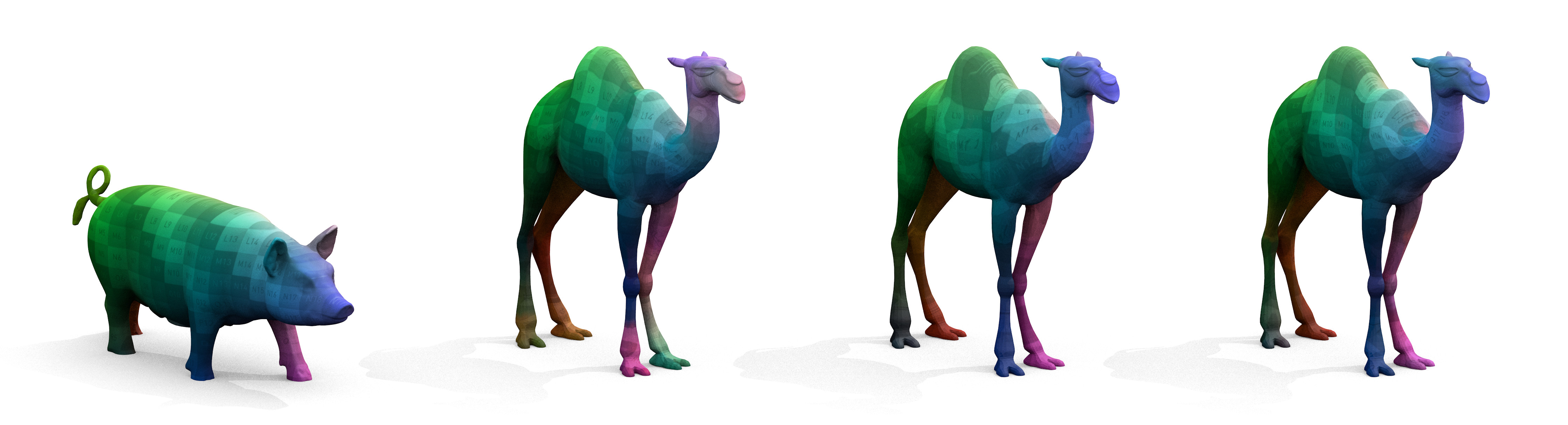}
        \put(10,27){Source\small}
        \put(32,27){Smooth Shells~\cite{eisenberger2020smooth}\small}
        \put(61,27){Ours\small}
        \put(83,27){Ours + SL\small}
\end{overpic}
\caption{\textbf{Unsupervised correspondences on SHREC20.} A qualitative comparison on non-isometric pairs from SHREC20~\cite{dyke2020shrec}.  While the correspondences predicted by our method are generally very accurate, the postprocessing still helps to remove local noise.
The baseline \cite{eisenberger2020smooth} naturally produces smooth matches, but global parts of the geometry are sometimes mismatched (compare for instance the front legs and and head of the camel shown here). 
}\label{fig:shrec20qual}
\end{figure*}


During training, we sample a pair of input shapes from our training set, predict an interpolation and a set of dense point-to-point correspondences and optimize the model parameters according to our composite loss~\eqref{eq:loss}.
All hyperparameters were selected on a validation set and the same configuration is used in all of our experiments.

Two parameters are varied during training: In the beginning, we set the number of discrete time steps to $T=1$ and then increase it on a logarithmic scale.
This multi-scale optimization strategy, which is motivated by classical non-learning interpolation algorithms~\cite{kilian2007geometric,heeren2012time}, leads to an overall faster and more robust convergence.
The geodesic loss $\ell_\mathrm{geo}$ initially helps to guide the optimization such that it converges to meaningful local minima.
On the other hand, we found that it can actually be detrimental in the case of extremely non-isometric pairs (e.g. two different classes of animals).
Therefore, we decay the weight $\lambda_\mathrm{geo}=0$ of this loss as a fine-tuning step during training after a fixed number of epochs.

As a form of data augmentation, we randomly subsample the triangulation of both input meshes separately and rotate the input pair along the azimuth axis in each iteration.
This prevents our method from relying on pairs with compatible connectivity, since we ideally want our predictions to be independent from the discretization.

At test time, we simply query our model to obtain an interpolation of an input pair of shapes.
The soft correspondences $\Pi$ obtained with our method are generally very accurate, but the conversion to hard correspondences (i.e. point-to-point matches) via thresholding leads to a certain degree of local noise.
To create more smooth correspondences, we additionally post-process our results with the multi-scale matching method smooth shells~\cite{eisenberger2020smooth}.
Post-processing is standard in unsupervised correspondence learning.

\begin{figure*}
\includegraphics[width=0.49\linewidth]{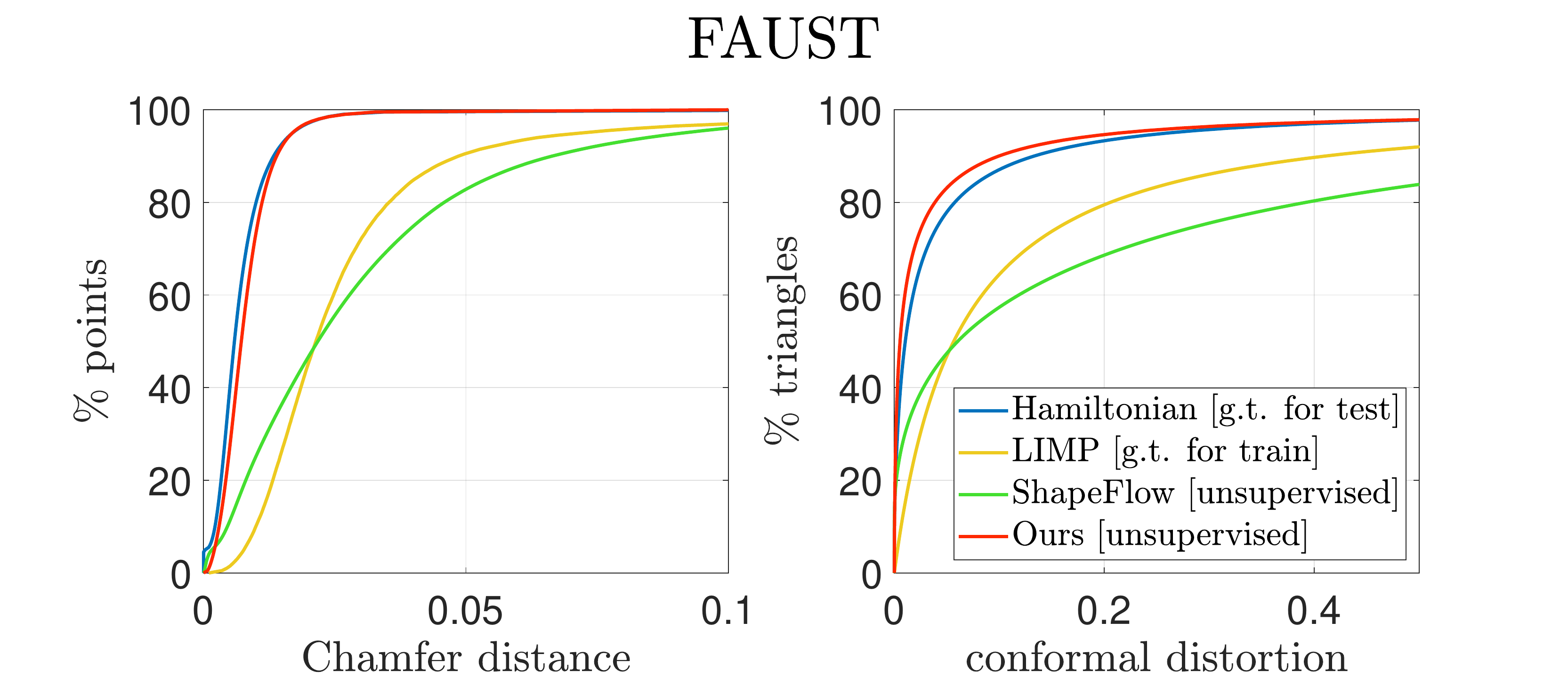}
\includegraphics[width=0.49\linewidth]{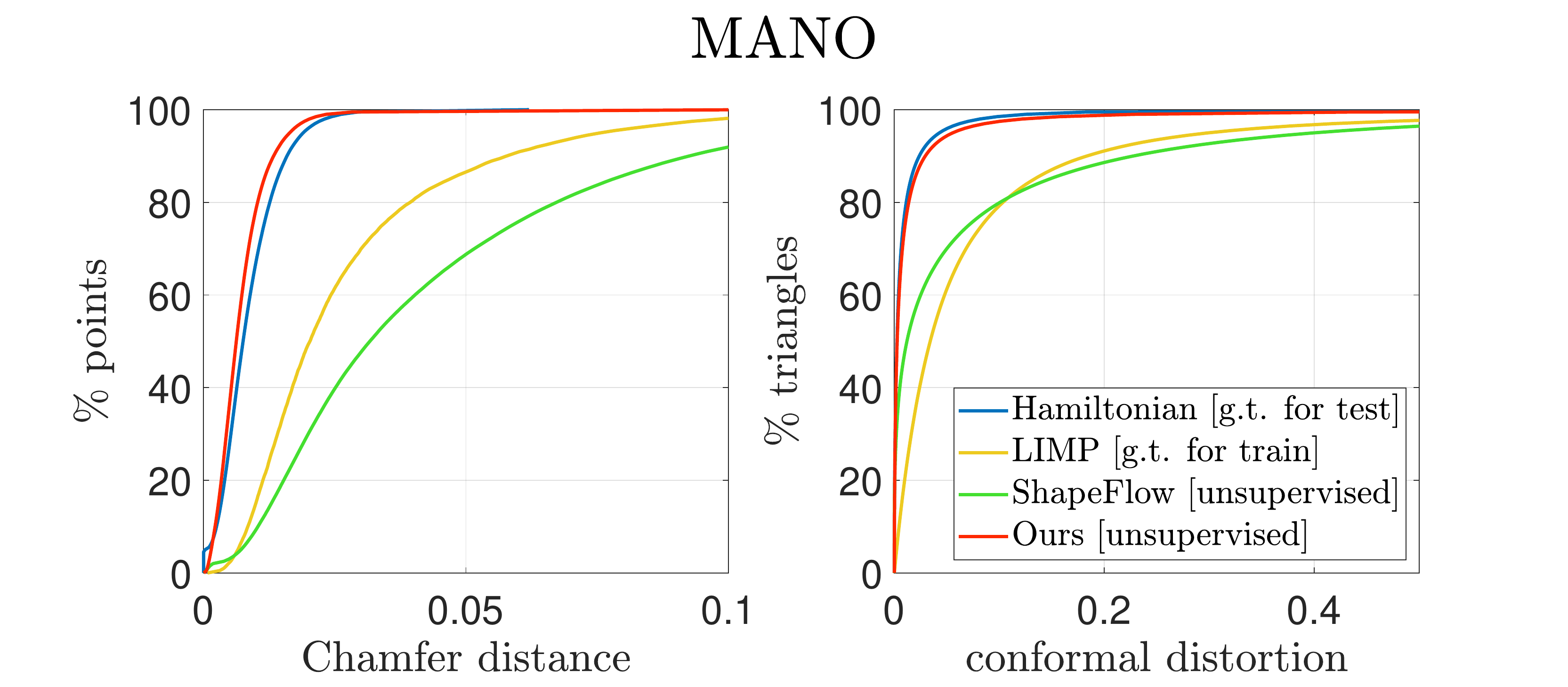}
\caption{\textbf{Interpolation on FAUST~\cite{Bogo:CVPR:2014} and MANO~\cite{MANO:SIGGRAPHASIA:2017}.} We show a quantitative comparison of interpolations obtained with Hamiltonian shape interpolation~\cite{eisenberger2020hamiltonian}, LIMP~\cite{cosmo2020limp}, ShapeFlow~\cite{jiang2020shapeflow} and our method. ShapeFlow~\cite{jiang2020shapeflow} computes an extrinsic flow to interpolate a pair of objects in an unsupervised manner, but they do not model shape correspondences explicitly which is suboptimal for the large pose variations of deformable object categories. On both benchmarks, our method outperforms LIMP~\cite{cosmo2020limp}, despite the fact that the it uses g.t.~correspondences for training.
It is also on par with the axiomatic baseline Hamiltonian interpolation~\cite{eisenberger2020hamiltonian}, which is remarkable since~\cite{eisenberger2020hamiltonian} requires dense correspondences even at test time.}\label{fig:manofaustcomparison}
\end{figure*}
\begin{figure}
\center
\begin{overpic}
        [width=0.95\linewidth,left]{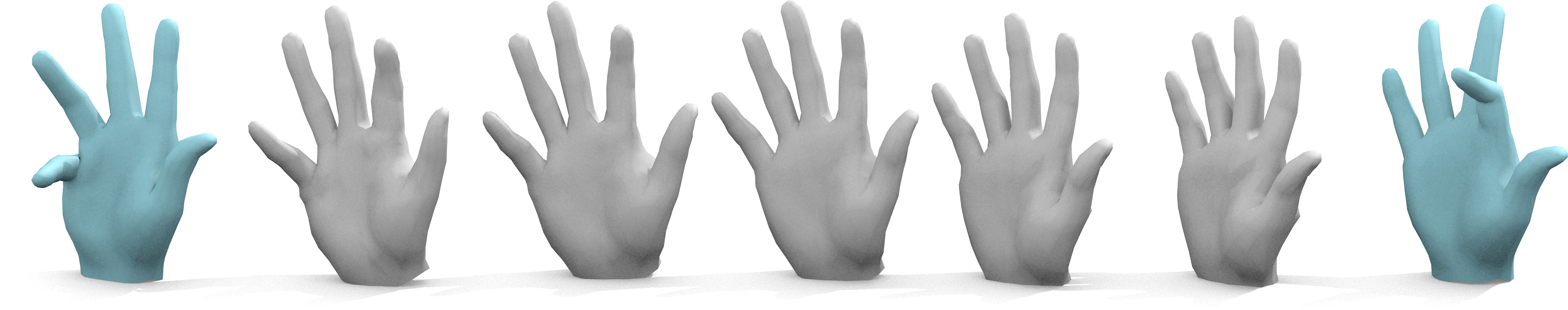}
        \put(95,14){\cite{cosmo2020limp}\small}
\end{overpic}
\begin{overpic}
        [width=0.95\linewidth,left]{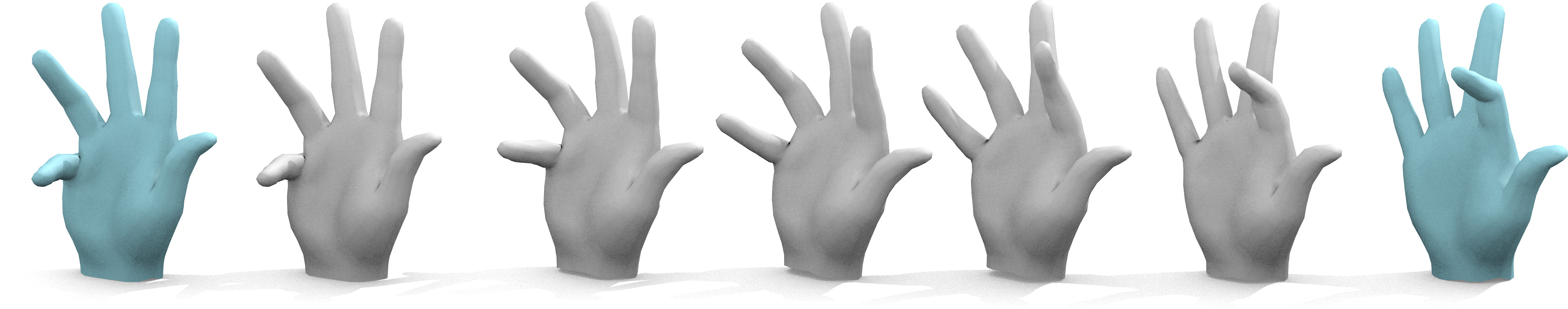}
        \put(94,14){\cite{eisenberger2020hamiltonian}\small}
\end{overpic}
\begin{overpic}
        [width=0.95\linewidth,left]{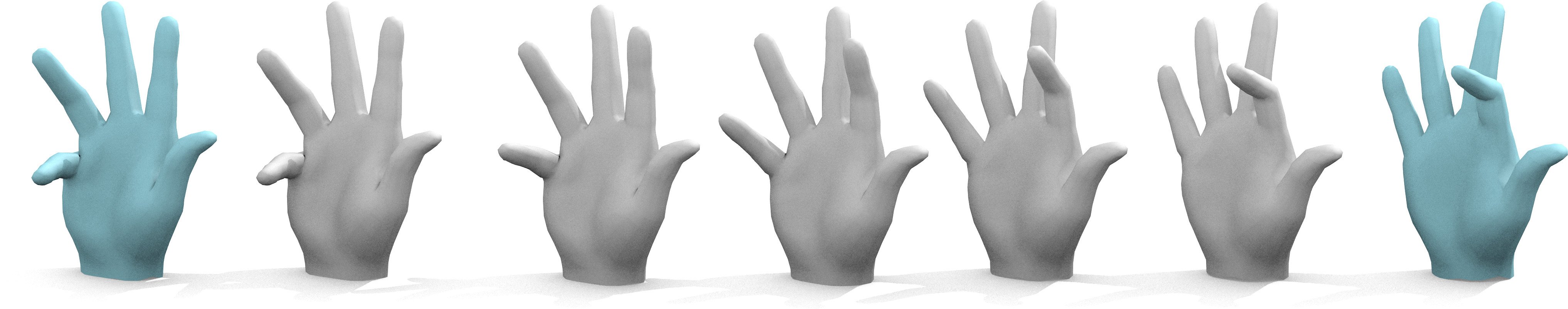}
        \put(92,14){Ours\small}
\end{overpic}
\caption{\textbf{Interpolation on MANO~\cite{MANO:SIGGRAPHASIA:2017}.}
We show the interpolation sequence (gray) for a pair (blue) from the test set. 
LIMP~\cite{cosmo2020limp}, which requires ground-truth correspondences for training, explicitly reconstructs the geometry of intermediate shapes in a variational autoencoder architecture which limits the generalization to unseen poses.
Hamiltonian shape interpolation~\cite{eisenberger2020hamiltonian} yields high quality results that are comparable to ours, but it is an axiomatic method that requires ground-truth correspondences at test time and multiple minutes of optimization per pair.}\label{fig:manoqual}
\end{figure}
\begin{figure*}
\centering
\begin{subfigure}{.21\linewidth}
  \centering
	\includegraphics[width=0.9\linewidth]{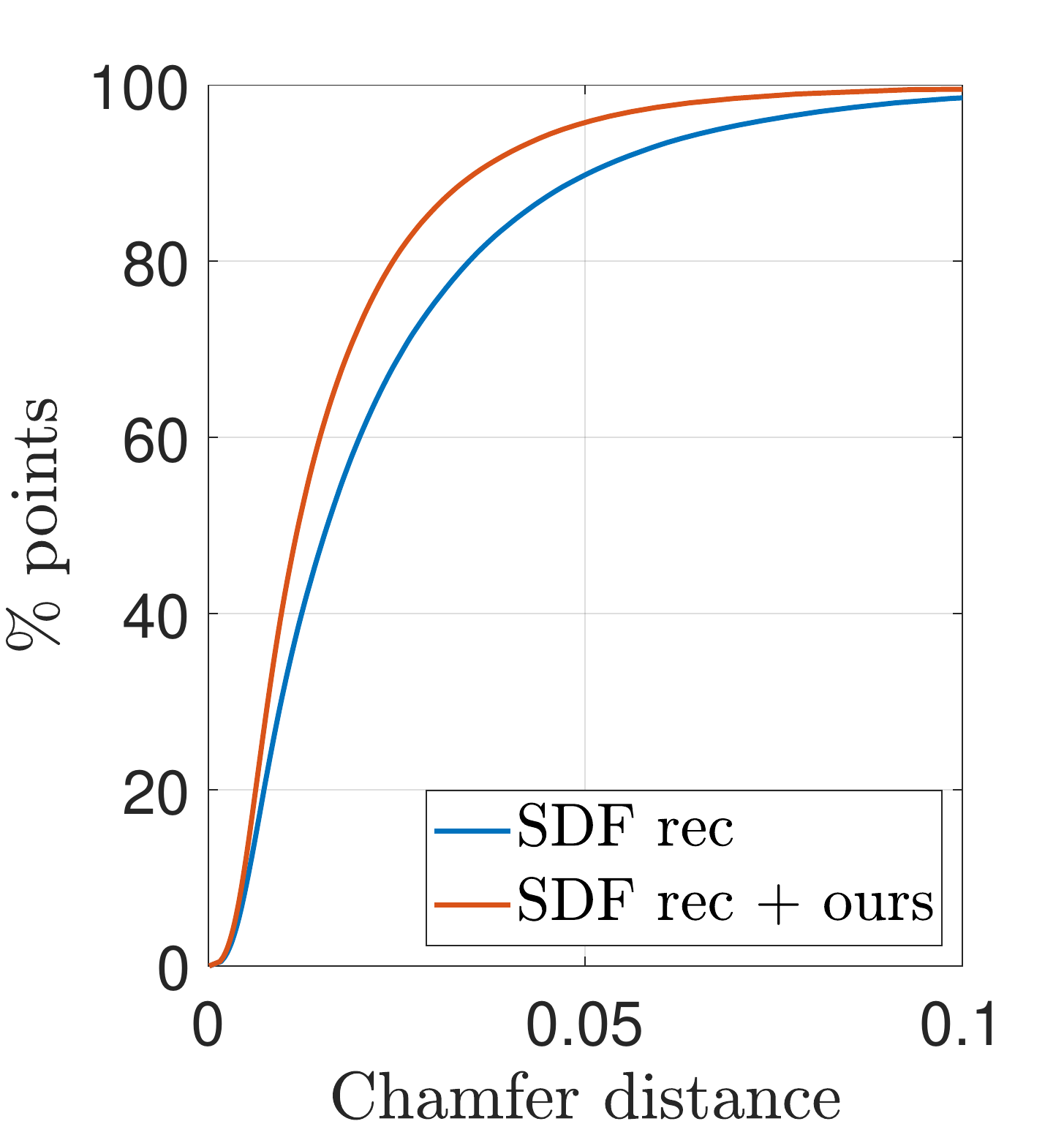}
\end{subfigure}
\begin{subfigure}{.78\linewidth}
\begin{overpic}
        [width=0.49\linewidth]{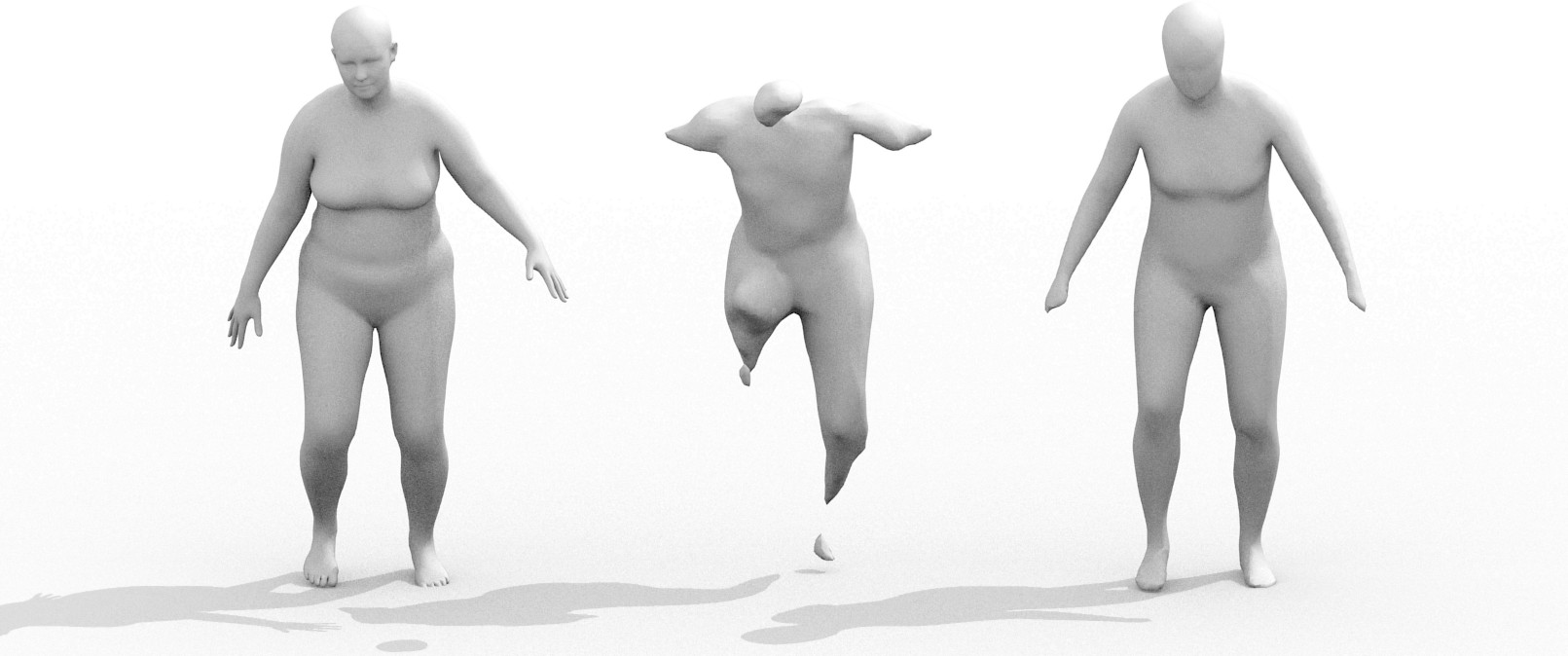}
        \put(18,47){Input\small}
        \put(45,47){\cite{icml2020_2086}\small}
        \put(67,47){\cite{icml2020_2086} + ours\small}
\end{overpic}
\begin{overpic}
        [width=0.49\linewidth]{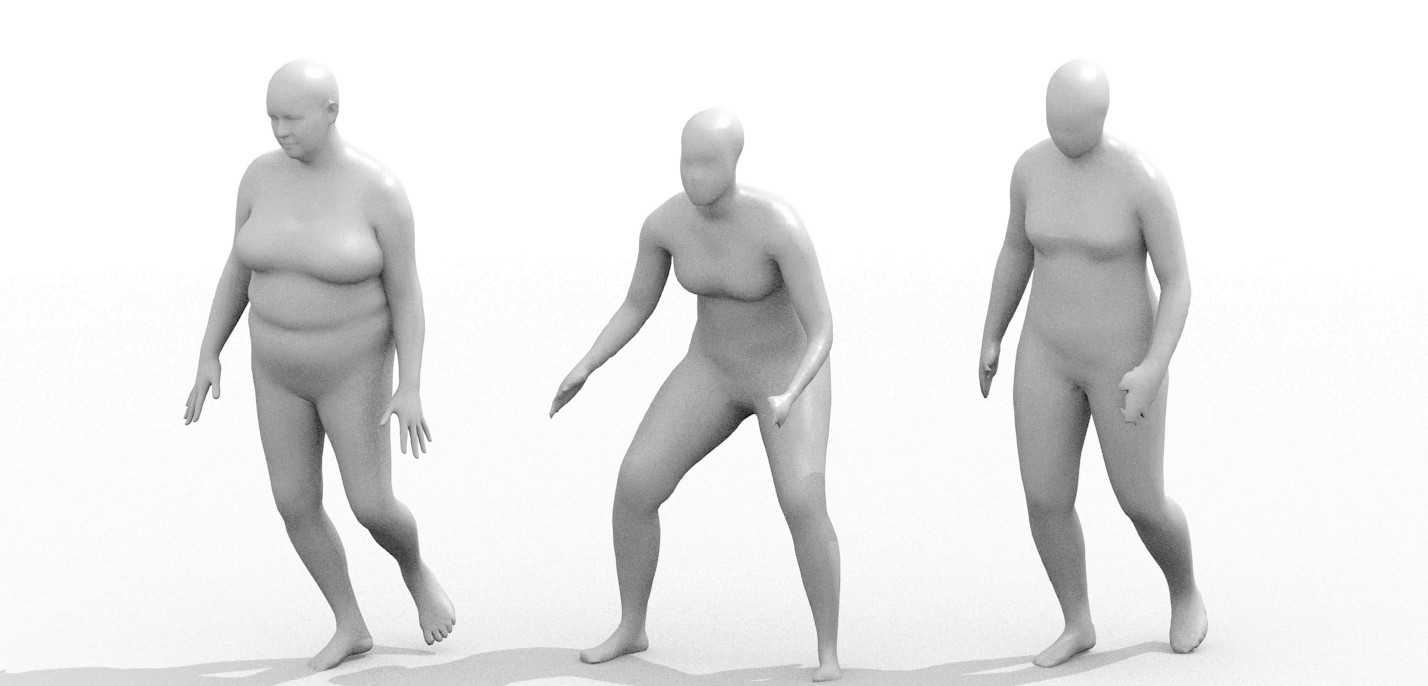}
        \put(18,47){Input\small}
        \put(45,47){\cite{icml2020_2086}\small}
        \put(67,47){\cite{icml2020_2086} + ours\small}
\end{overpic}
  \centering
\end{subfigure}
\caption{\textbf{Data augmentation for implicit surface reconstruction.} We show, as a proof of concept, that our method can be used to augment a small training set with additional poses. We verify this by comparing the reconstruction error of an implicit surface reconstruction method~\cite{icml2020_2086} with and without data augmentation. The general idea behind this is to construct a shape space that encodes arbitrary poses in a latent representation. As we show here, supplementing collections of sparse observations with intermediate poses constitutes a natural extension which helps to learn a meaningful shape distribution.\vspace*{-5pt}}\label{fig:sdfrec}
\end{figure*}

\section{Experiments}\label{s:exp}

We now evaluate the performance of \method in terms of shape correspondence and interpolation (Sec.~\ref{subsec:shapecorr},~\ref{subsec:shapeipol}), as well as for data augmentation in~Sec.~\ref{subsec:sdfrec}.


\subsection{Shape correspondence}\label{subsec:shapecorr}

\paragraph{Datasets.}

We evaluate the matching accuracy of our method on two benchmarks.
The first is FAUST~\cite{Bogo:CVPR:2014}, which contains 10 humans with 10 different poses each.
We split it in a training and test set of 80 and 20 shapes respectively.
Instead of the standard meshes, we use the more recent version of the benchmark~\cite{ren2018orientation} where each shape was re-meshed individually.
This makes it challenging but also more realistic, since for real-world scans the sampling of surfaces is generally incompatible.
The second benchmark we consider is the recent SHREC20 challenge~\cite{dyke2020shrec} which focuses on non-isometric deformations.
It contains 14 shapes of different animals, some of which are real scans with holes, topological changes and partial geometries.
The ground truth for this dataset consists of sparse annotated keypoints which we use for evaluation.
Since there are no dense annotated point-to-point correspondences, most existing supervised methods do not apply here.
The final benchmark we show is G-S-H (Galgo, Sphynx, Human), for which we created our own dataset, see~\Cref{s:galgo} for more details. It contains non-isometric pairs from three object categories (a dog 'Galgo', a cat 'Sphynx' and a human) with multiple challenging poses each, as well as dense ground truth matches.

\paragraph{Evaluation metrics.}

Following the Princeton benchmark protocol~\cite{kim11}, the accuracy of a set of point-to-point correspondences is defined as the geodesic distance of the predicted and the ground-truth matches, normalized by the square root area of the mesh.
For FAUST remeshed, we compute the distance for all points, whereas for SHREC20 this is done for all available sparse annotations.

\paragraph{Discussion.}

As shown in~\Cref{table:matchingaccuracy},~\Cref{fig:shrec20comparison} and~\Cref{fig:heatmaps}, \method obtains state-of-the-art results on FAUST remeshed, SHREC20 and our own benchmark G-S-H, respectively.
The overall suboptimal performance of existing methods on the latter two benchmarks can be attributed to the fact that, expect for~\cite{eisenberger2020smooth}, most of them implicitly assume near-isometry or at least compatible local features.
This, however, does not hold for most examples in SHREC20 (see~\Cref{fig:shrec20qual} for a qualitative comparison) and G-S-H  (for instance, on SHREC20, \method matches 92\% of the vertices within 0.25 geodesic error vs.~79\% of the second best, smooth shells).
\method is particularly good at discovering structural correspondences, which can then be further refined in post-processing.

\subsection{Shape interpolation}\label{subsec:shapeipol}

\paragraph{Datasets.}

For shape interpolation, we report results on the FAUST~\cite{Bogo:CVPR:2014} (see~\Cref{subsec:shapecorr}) and MANO~\cite{MANO:SIGGRAPHASIA:2017} datasets.
The latter consists of synthetic hands in various poses --- we use $100$ shapes for training and $20$ different samples for testing.

\paragraph{Evaluation metrics.}

We use two metrics to quantify the precision of an interpolation.
The conformal distortion metric signifies how much individual triangles of a mesh distort throughout an interpolation sequence, in comparison to the reference pose $\mathcal{X}$, see~\cite[Eq.~(3)]{hormann2000mips} for a definition.
Less distortion corresponds to more realistic shapes.
The other metric we consider measures the reconstruction error of the target shape $\mathcal{Y}$, defined as the Chamfer distance between $\mathcal{Y}$ and the deformed shape $\mathcal{X}(1)$.
A good overlap at $t=1$ is an important quality criterion because, while our interpolations exactly coincide with the first shape $\mathcal{X}=\mathcal{X}(0)$, they only approximately align with $\mathcal{Y}\approx\mathcal{X}(1)$.
The same holds true for the three baselines~\cite{cosmo2020limp,jiang2020shapeflow,eisenberger2020hamiltonian} that we compare against.

\paragraph{Discussion.}

Results are shown in~\Cref{fig:manofaustcomparison}.
On both of these benchmarks, our method significantly outperforms the supervised baseline LIMP~\cite{cosmo2020limp} which requires ground-truth correspondences for training. Similar to our approach, the unsupervised method ShapeFlow~\cite{jiang2020shapeflow} continuously deforms a given input shape to obtain an interpolation. However, they do not estimate correspondences explicitly which limits the performance on deformable object categories like humans, animals or hands\footnote{These types of objects typically have a high pose variation with large degrees of non-rigid deformations, whereas ShapeFlow mainly specializes on man-made objects like chairs or cars. The few deformable examples that they show in~\cite[fig. 5]{jiang2020shapeflow} are intended as a proof of concept since they use ground-truth correspondences and overfit on a single pair of shapes.}.
More surprisingly, our approach is even on par with the axiomatic, non-learning interpolation baseline~\cite{eisenberger2020hamiltonian} which requires to know dense correspondences at \emph{test} time.
See~\Cref{fig:manoqual} for a qualitative comparison on MANO.

\subsection{Application: data augmentation}\label{subsec:sdfrec}

Our method is, to the best of our knowledge, the first one that jointly predicts correspondences and an interpolation of deformable objects in a single learning framework.
As an application of unsupervised interpolation, we show how our method can be used to create additional training samples as a form of data augmentation.
To that end, we train an implicit surface reconstruction method~\cite{icml2020_2086} on a small set of $20$ SMPL shapes from the SURREAL dataset~\cite{varol2017learning} and evaluate the obtained reconstructions on a separate test set of $100$ shapes.
Additionally, we use our method to create 3 additional, interpolated training poses for each pair in the training set and compare the results with the vanilla training.
To measure the quality of the obtained reconstructions, we report the reconstruction error on the test set, defined as the Chamfer distance of the test shapes to the reconstructed surface, see~\Cref{fig:sdfrec}.

Overall, these results indicate that using our method to enlarge a training set of 3D shapes can be useful for downstream tasks, especially when training data is limited.

\section{Conclusions}\label{s:conclusions}

We presented a new framework for 3D shape understanding that simultaneously addresses the problems of shape correspondence and interpolation.
The key insight we want to advocate is that these two goals mutually reinforce each other: Better correspondences yield more accurate interpolations and, vice versa, meaningful deformations of 3D surfaces act as a strong geometric prior for finding correspondences.
In comparison to related existing approaches, our model can be trained in a fully unsupervised manner and generates correspondence and interpolation in a single pass.
We show that our method produces stable results for a variety of correspondence and interpolation tasks, including challenging inter-class pairs with high degrees of non-isometric deformations.
We expect that NeuroMorph will facilitate 3D shape analysis on large real-world datasets where obtaining exact ground-truth matches is prohibitively expensive.

\section*{Acknowledgements}

We would like to thank Matan Atzmon and Aysim Toker for useful discussions and Roberto Dyke for quick help with the SHREC20 ground truth.
This work was supported by the Munich Center for Machine Learning and by the ERC Advanced Grant SIMULACRON.

{\small\bibliographystyle{ieee_fullname}
\bibliography{ms}}

\begin{thebibliography}{10}\itemsep=-1pt

\bibitem{aflalo2016spectral}
Yonathan Aflalo, Anastasia Dubrovina, and Ron Kimmel.
\newblock Spectral generalized multi-dimensional scaling.
\newblock {\em IJCV}, 118(3):380--392, 2016.

\bibitem{Bogo:CVPR:2014}
Federica Bogo, Javier Romero, Matthew Loper, and Michael~J. Black.
\newblock {FAUST}: Dataset and evaluation for {3D} mesh registration.
\newblock In {\em Proceedings IEEE Conf. on Computer Vision and Pattern
  Recognition (CVPR)}, Piscataway, NJ, USA, June 2014. IEEE.

\bibitem{boscaini2016learning}
Davide Boscaini, Jonathan Masci, Emanuele Rodol{\`a}, and Michael Bronstein.
\newblock Learning shape correspondence with anisotropic convolutional neural
  networks.
\newblock In {\em Advances in neural information processing systems}, pages
  3189--3197, 2016.

\bibitem{bronstein2006generalized}
Alexander~M Bronstein, Michael~M Bronstein, and Ron Kimmel.
\newblock Generalized multidimensional scaling: a framework for
  isometry-invariant partial surface matching.
\newblock {\em PNAS}, 103(5):1168--1172, 2006.

\bibitem{bronstein2017geometric}
Michael~M Bronstein, Joan Bruna, Yann LeCun, Arthur Szlam, and Pierre
  Vandergheynst.
\newblock Geometric deep learning: going beyond euclidean data.
\newblock {\em IEEE Signal Processing Magazine}, 34(4):18--42, 2017.

\bibitem{choy20163d}
Christopher~B Choy, Danfei Xu, JunYoung Gwak, Kevin Chen, and Silvio Savarese.
\newblock 3d-r2n2: A unified approach for single and multi-view 3d object
  reconstruction.
\newblock In {\em European conference on computer vision}, pages 628--644.
  Springer, 2016.

\bibitem{cosmo2020limp}
Luca Cosmo, Antonio Norelli, Oshri Halimi, Ron Kimmel, and Emanuele Rodol{\`a}.
\newblock Limp: Learning latent shape representations with metric preservation
  priors.
\newblock {\em arXiv preprint arXiv:2003.12283}, 2020.

\bibitem{defferrard2016convolutional}
Micha{\"e}l Defferrard, Xavier Bresson, and Pierre Vandergheynst.
\newblock Convolutional neural networks on graphs with fast localized spectral
  filtering.
\newblock In {\em Advances in neural information processing systems}, pages
  3844--3852, 2016.

\bibitem{donati2020deep}
Nicolas Donati, Abhishek Sharma, and Maks Ovsjanikov.
\newblock Deep geometric functional maps: Robust feature learning for shape
  correspondence.
\newblock {\em arXiv preprint arXiv:2003.14286}, 2020.

\bibitem{dyke2020shrec}
Roberto~M Dyke, Yu-Kun Lai, Paul~L Rosin, Stefano Zappal{\`a}, Seana Dykes,
  Daoliang Guo, Kun Li, Riccardo Marin, Simone Melzi, and Jingyu Yang.
\newblock Shrec’20: Shape correspondence with non-isometric deformations.
\newblock {\em Computers \& Graphics}, 92:28--43, 2020.

\bibitem{eisenberger2020hamiltonian}
Marvin Eisenberger and Daniel Cremers.
\newblock Hamiltonian dynamics for real-world shape interpolation.
\newblock In {\em ECCV}, 2020.

\bibitem{eisenberger2019divergence}
Marvin Eisenberger, Zorah L{\"a}hner, and Daniel Cremers.
\newblock Divergence-free shape correspondence by deformation.
\newblock In {\em Computer Graphics Forum}, volume~38, pages 1--12. Wiley
  Online Library, 2019.

\bibitem{eisenberger2020smooth}
Marvin Eisenberger, Zorah Lahner, and Daniel Cremers.
\newblock Smooth shells: Multi-scale shape registration with functional maps.
\newblock In {\em Proceedings of the IEEE/CVF Conference on Computer Vision and
  Pattern Recognition}, pages 12265--12274, 2020.

\bibitem{eisenberger2020deep}
Marvin Eisenberger, Aysim Toker, Laura Leal-Taixe, and Daniel Cremers.
\newblock Deep shells: Unsupervised shape correspondence with optimal
  transport.
\newblock {\em arXiv preprint}, 2020.

\bibitem{fan2017point}
Haoqiang Fan, Hao Su, and Leonidas~J Guibas.
\newblock A point set generation network for 3d object reconstruction from a
  single image.
\newblock In {\em Proceedings of the IEEE conference on computer vision and
  pattern recognition}, pages 605--613, 2017.

\bibitem{gkioxari2019mesh}
Georgia Gkioxari, Jitendra Malik, and Justin Johnson.
\newblock Mesh r-cnn.
\newblock In {\em Proceedings of the IEEE International Conference on Computer
  Vision}, pages 9785--9795, 2019.

\bibitem{icml2020_2086}
Amos Gropp, Lior Yariv, Niv Haim, Matan Atzmon, and Yaron Lipman.
\newblock Implicit geometric regularization for learning shapes.
\newblock In {\em Proceedings of Machine Learning and Systems 2020}, pages
  3569--3579. 2020.

\bibitem{groueix20183dcoded}
Thibault Groueix, Matthew Fisher, Vladimir~G. Kim, Bryan~C. Russell, and
  Mathieu Aubry.
\newblock 3d-coded: 3d correspondences by deep deformation.
\newblock In {\em The European Conference on Computer Vision (ECCV)}, September
  2018.

\bibitem{groueix2018papier}
Thibault Groueix, Matthew Fisher, Vladimir~G Kim, Bryan~C Russell, and Mathieu
  Aubry.
\newblock A papier-mache approach to learning 3d surface generation.
\newblock In {\em Proceedings of the IEEE conference on computer vision and
  pattern recognition}, pages 216--224, 2018.

\bibitem{halimi2019unsupervised}
Oshri Halimi, Or Litany, Emanuele Rodola, Alex~M Bronstein, and Ron Kimmel.
\newblock Unsupervised learning of dense shape correspondence.
\newblock In {\em Proceedings of the IEEE Conference on Computer Vision and
  Pattern Recognition}, pages 4370--4379, 2019.

\bibitem{he15deep}
Kaiming He, Xiangyu Zhang, Shaoqing Ren, and Jian Sun.
\newblock Deep residual learning for image recognition.
\newblock {\em arXiv preprint arXiv:1512.03385}, 2015.

\bibitem{heeren2016shellsplines}
Behrend Heeren, Martin Rumpf, Peter Schr\"oder, Max Wardetzky, and Benedikt
  Wirth.
\newblock Splines in the space of shells.
\newblock {\em Computer Graphics Forum}, 35(5):111--120, 2016.

\bibitem{heeren2012time}
Behrend Heeren, Martin Rumpf, Max Wardetzky, and Benedikt Wirth.
\newblock Time-discrete geodesics in the space of shells.
\newblock In {\em Computer Graphics Forum}, volume~31, pages 1755--1764. Wiley
  Online Library, 2012.

\bibitem{heeren2018principal}
Behrend Heeren, Chao Zhang, Martin Rumpf, and William Smith.
\newblock Principal geodesic analysis in the space of discrete shells.
\newblock In {\em Computer Graphics Forum}, volume~37, pages 173--184. Wiley
  Online Library, 2018.

\bibitem{hormann2000mips}
Kai Hormann and G{\"u}nther Greiner.
\newblock Mips: An efficient global parametrization method.
\newblock Technical report, Erlangen-Nuernberg University (Germany) Computer
  Graphics Group, 2000.

\bibitem{jiang2020shapeflow}
Chiyu Jiang, Jingwei Huang, Andrea Tagliasacchi, Leonidas Guibas, et~al.
\newblock Shapeflow: Learnable deformations among 3d shapes.
\newblock {\em arXiv preprint arXiv:2006.07982}, 2020.

\bibitem{kilian2007geometric}
Martin Kilian, Niloy~J Mitra, and Helmut Pottmann.
\newblock Geometric modeling in shape space.
\newblock In {\em ACM Transactions on Graphics (TOG)}, volume~26, page~64. ACM,
  2007.

\bibitem{kim11}
Vladimir~G Kim, Yaron Lipman, and Thomas~A Funkhouser.
\newblock Blended intrinsic maps.
\newblock {\em Transactions on Graphics {(TOG)}}, 30(4), 2011.

\bibitem{kipf2016semi}
Thomas~N Kipf and Max Welling.
\newblock Semi-supervised classification with graph convolutional networks.
\newblock {\em arXiv preprint arXiv:1609.02907}, 2016.

\bibitem{litany2017deep}
Or Litany, Tal Remez, Emanuele Rodol{\`a}, Alex Bronstein, and Michael
  Bronstein.
\newblock Deep functional maps: Structured prediction for dense shape
  correspondence.
\newblock In {\em Proceedings of the IEEE International Conference on Computer
  Vision}, pages 5659--5667, 2017.

\bibitem{litany2017fullyspectral}
Or Litany, Emanuele Rodol{\`a}, Alex Bronstein, and Michael Bronstein.
\newblock Fully spectral partial shape matching.
\newblock {\em Computer Graphics Forum}, 36(2):1681--1707, 2017.

\bibitem{litany2016puzzles}
Or Litany, Emanuele Rodol\`a, Alex~M Bronstein, Michael~M Bronstein, and Daniel
  Cremers.
\newblock Non-rigid puzzles.
\newblock {\em {Computer Graphics Forum (CGF), Proceedings of Symposium on
  Geometry Processing (SGP)}}, 35(5), 2016.

\bibitem{loper15smpl:}
Matthew Loper, Naureen Mahmood, Javier Romero, Gerard Pons-Moll, and Michael~J.
  Black.
\newblock {SMPL}: a skinned multi-person linear model.
\newblock {\em ACM Trans. on Graphics (TOG)}, 2015.

\bibitem{masci2015geodesic}
Jonathan Masci, Davide Boscaini, Michael Bronstein, and Pierre Vandergheynst.
\newblock Geodesic convolutional neural networks on riemannian manifolds.
\newblock In {\em Proceedings of the IEEE international conference on computer
  vision workshops}, pages 37--45, 2015.

\bibitem{melzi2019zoomout}
Simone Melzi, Jing Ren, Emanuele Rodol{\`a}, Abhishek Sharma, Peter Wonka, and
  Maks Ovsjanikov.
\newblock Zoomout: Spectral upsampling for efficient shape correspondence.
\newblock {\em ACM Transactions on Graphics (TOG)}, 38(6):155, 2019.

\bibitem{mescheder2019occupancy}
Lars Mescheder, Michael Oechsle, Michael Niemeyer, Sebastian Nowozin, and
  Andreas Geiger.
\newblock Occupancy networks: Learning 3d reconstruction in function space.
\newblock In {\em Proceedings of the IEEE Conference on Computer Vision and
  Pattern Recognition}, pages 4460--4470, 2019.

\bibitem{monti2017geometric}
Federico Monti, Davide Boscaini, Jonathan Masci, Emanuele Rodola, Jan Svoboda,
  and Michael~M Bronstein.
\newblock Geometric deep learning on graphs and manifolds using mixture model
  cnns.
\newblock In {\em Proceedings of the IEEE Conference on Computer Vision and
  Pattern Recognition}, pages 5115--5124, 2017.

\bibitem{niemeyer2019occupancy}
Michael Niemeyer, Lars Mescheder, Michael Oechsle, and Andreas Geiger.
\newblock Occupancy flow: 4d reconstruction by learning particle dynamics.
\newblock In {\em Proceedings of the IEEE International Conference on Computer
  Vision}, pages 5379--5389, 2019.

\bibitem{ovsjanikov2012functional}
Maks Ovsjanikov, Mirela Ben-Chen, Justin Solomon, Adrian Butscher, and Leonidas
  Guibas.
\newblock Functional maps: a flexible representation of maps between shapes.
\newblock {\em ACM Transactions on Graphics (TOG)}, 31(4):30, 2012.

\bibitem{park2019deepsdf}
Jeong~Joon Park, Peter Florence, Julian Straub, Richard Newcombe, and Steven
  Lovegrove.
\newblock Deepsdf: Learning continuous signed distance functions for shape
  representation.
\newblock In {\em Proceedings of the IEEE Conference on Computer Vision and
  Pattern Recognition}, pages 165--174, 2019.

\bibitem{poulenard2018multi}
Adrien Poulenard and Maks Ovsjanikov.
\newblock Multi-directional geodesic neural networks via equivariant
  convolution.
\newblock {\em ACM Transactions on Graphics (TOG)}, 37(6):1--14, 2018.

\bibitem{qi2017pointnet}
Charles~R Qi, Hao Su, Kaichun Mo, and Leonidas~J Guibas.
\newblock Pointnet: Deep learning on point sets for 3d classification and
  segmentation.
\newblock In {\em Proceedings of the IEEE conference on computer vision and
  pattern recognition}, pages 652--660, 2017.

\bibitem{qi2017pointnet++}
Charles~Ruizhongtai Qi, Li Yi, Hao Su, and Leonidas~J Guibas.
\newblock Pointnet++: Deep hierarchical feature learning on point sets in a
  metric space.
\newblock In {\em Advances in neural information processing systems}, pages
  5099--5108, 2017.

\bibitem{ren2018orientation}
Jing Ren, Adrien Poulenard, Peter Wonka, and Maks Ovsjanikov.
\newblock Continuous and orientation-preserving correspondences via functional
  maps.
\newblock {\em ACM Trans. Graph.}, 37(6):248:1--248:16, Dec. 2018.

\bibitem{rodola2016partial}
Emanuele Rodolà, Luca Cosmo, Michael Bronstein, Andrea Torsello, and Daniel
  Cremers.
\newblock Partial functional correspondence.
\newblock {\em Computer Graphics Forum (CGF)}, 2016.

\bibitem{MANO:SIGGRAPHASIA:2017}
Javier Romero, Dimitrios Tzionas, and Michael~J. Black.
\newblock Embodied hands: Modeling and capturing hands and bodies together.
\newblock {\em ACM Transactions on Graphics, (Proc. SIGGRAPH Asia)}, 36(6),
  Nov. 2017.

\bibitem{roufosse2019unsupervised}
Jean-Michel Roufosse, Abhishek Sharma, and Maks Ovsjanikov.
\newblock Unsupervised deep learning for structured shape matching.
\newblock In {\em Proceedings of the IEEE International Conference on Computer
  Vision}, pages 1617--1627, 2019.

\bibitem{Wiersma2020}
Klaus~Hildebrandt Ruben~Wiersma, Elmar~Eisemann.
\newblock Cnns on surfaces using rotation-equivariant features.
\newblock {\em Transactions on Graphics}, 39(4), July 2020.

\bibitem{sahilliouglu2019recent}
Yusuf Sahillio{\u{g}}lu.
\newblock Recent advances in shape correspondence.
\newblock {\em The Visual Computer}, pages 1--17, 2019.

\bibitem{sharma2020weakly}
Abhishek Sharma and Maks Ovsjanikov.
\newblock Weakly supervised deep functional map for shape matching.
\newblock {\em arXiv preprint arXiv:2009.13339}, 2020.

\bibitem{sorkine2007rigid}
Olga Sorkine and Marc Alexa.
\newblock As-rigid-as-possible surface modeling.
\newblock In {\em Symposium on Geometry processing}, volume~4, pages 109--116,
  2007.

\bibitem{tam2013pointcloudsurvey}
Gary K.~L. Tam, Zhi-Quan Cheng, Yu-Kun Lai, Frank~C. Langbein, Yonghuai Liu,
  David Marshall, Ralph~R. Martin, Xian-Fang Sun, and Paul~L. Rosin.
\newblock Registration of 3d point clouds and meshes: A survey from rigid to
  nonrigid.
\newblock {\em IEEE Transactions on Visualization and Computer Graphics},
  19(7):1199--1217, July 2013.

\bibitem{thomas2019kpconv}
Hugues Thomas, Charles~R Qi, Jean-Emmanuel Deschaud, Beatriz Marcotegui,
  Fran{\c{c}}ois Goulette, and Leonidas~J Guibas.
\newblock Kpconv: Flexible and deformable convolution for point clouds.
\newblock In {\em Proceedings of the IEEE International Conference on Computer
  Vision}, pages 6411--6420, 2019.

\bibitem{tombari2010SHOT}
Federico Tombari, Samuele Salti, and Luigi Di~Stefano.
\newblock Unique signatures of histograms for local surface description.
\newblock {\em In Proceedings of European Conference on Computer Vision
  {(ECCV)}}, 16(9):356--369, 2010.

\bibitem{vankaick11correspsurvey}
Oliver van Kaick, Hao Zhang, Ghassan Hamarneh, and Daniel Cohen-Or.
\newblock A survey on shape correspondence.
\newblock {\em Computer Graphics Forum}, 30(6):1681--1707, 2011.

\bibitem{varol2017learning}
Gul Varol, Javier Romero, Xavier Martin, Naureen Mahmood, Michael~J Black, Ivan
  Laptev, and Cordelia Schmid.
\newblock Learning from synthetic humans.
\newblock In {\em Proceedings of the IEEE Conference on Computer Vision and
  Pattern Recognition}, pages 109--117, 2017.

\bibitem{vestner2017pmf}
Matthias Vestner, Roee Litman, Emanuele Rodol\`a, Alex~M Bronstein, and Daniel
  Cremers.
\newblock Product manifold filter: Non-rigid shape correspondence via kernel
  density estimation in the product space.
\newblock In {\em IEEE Conference on Computer Vision and Pattern Recognition
  (CVPR)}, 2017.

\bibitem{von2016optimized}
Philipp von Radziewsky, Elmar Eisemann, Hans-Peter Seidel, and Klaus
  Hildebrandt.
\newblock Optimized subspaces for deformation-based modeling and shape
  interpolation.
\newblock {\em Computers \& Graphics}, 58:128--138, 2016.

\bibitem{wang2011discrete}
Chaohui Wang, Michael~M Bronstein, Alexander~M Bronstein, and Nikos Paragios.
\newblock Discrete minimum distortion correspondence problems for non-rigid
  shape matching.
\newblock In {\em International Conference on Scale Space and Variational
  Methods in Computer Vision}, pages 580--591. Springer, 2011.

\bibitem{wang20193dn}
Weiyue Wang, Duygu Ceylan, Radomir Mech, and Ulrich Neumann.
\newblock 3dn: 3d deformation network.
\newblock In {\em Proceedings of the IEEE Conference on Computer Vision and
  Pattern Recognition}, pages 1038--1046, 2019.

\bibitem{wang2019dynamic}
Yue Wang, Yongbin Sun, Ziwei Liu, Sanjay~E Sarma, Michael~M Bronstein, and
  Justin~M Solomon.
\newblock Dynamic graph cnn for learning on point clouds.
\newblock {\em Acm Transactions On Graphics (tog)}, 38(5):1--12, 2019.

\bibitem{wirth2011shapespace}
Benedikt Wirth, Leah Bar, Martin Rumpf, and Guillermo Sapiro.
\newblock A continuum mechanical approach to geodesics in shape space.
\newblock {\em International Journal of Computer Vision}, 93(3):293--318, Jul
  2011.

\bibitem{yang2015go}
Jiaolong Yang, Hongdong Li, Dylan Campbell, and Yunde Jia.
\newblock Go-icp: A globally optimal solution to 3d icp point-set registration.
\newblock {\em IEEE transactions on pattern analysis and machine intelligence},
  38(11):2241--2254, 2015.

\bibitem{yang2018foldingnet}
Yaoqing Yang, Chen Feng, Yiru Shen, and Dong Tian.
\newblock Foldingnet: Point cloud auto-encoder via deep grid deformation.
\newblock In {\em Proceedings of the IEEE Conference on Computer Vision and
  Pattern Recognition}, pages 206--215, 2018.

\bibitem{zhang2015shell}
Chao Zhang, Behrend Heeren, Martin Rumpf, and William~AP Smith.
\newblock Shell pca: Statistical shape modelling in shell space.
\newblock In {\em Proceedings of the IEEE International Conference on Computer
  Vision}, pages 1671--1679, 2015.

\bibitem{zhou2016fastglobal}
Qian-Yi Zhou, Jaesik Park, and Vladlen Koltun.
\newblock Fast global registration.
\newblock {\em Proceedings of European Conference on Computer Vision (ECCV)},
  2016.

\end{thebibliography}

\clearpage

\appendix

\section{Our dataset G-S-H}\label{s:galgo}

\begin{figure*}
\centering
\includegraphics[width=0.99\linewidth]{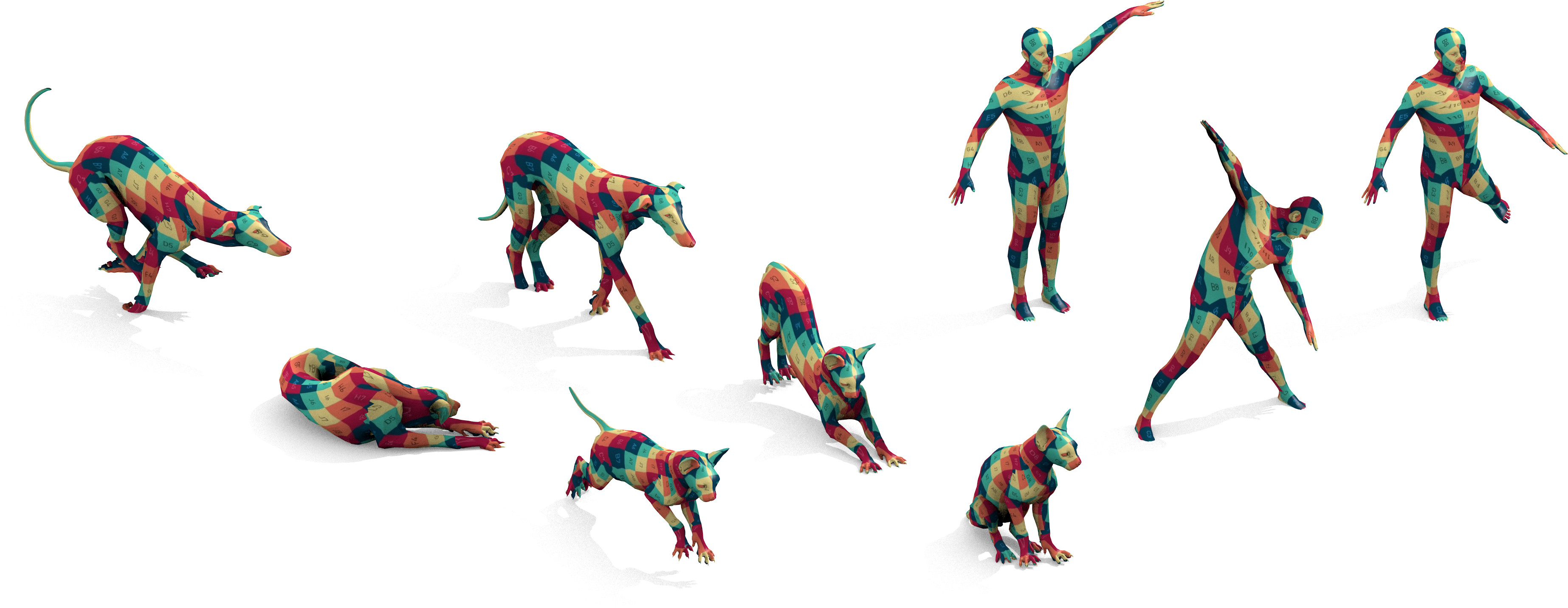}
\caption{\textbf{The G-S-H dataset.} We show 3 examples each for the 3 classes in our G-S-H dataset. Note, that all 3 classes share the same parameterization, despite the varying body proportions. In particular this means that we can obtain dense ground truth correspondences between all pairs of shapes, which we indicate here with a texture map.
}\label{fig:gsh3x3}
\end{figure*}

\begin{figure*}
\centering
\includegraphics[width=0.99\linewidth]{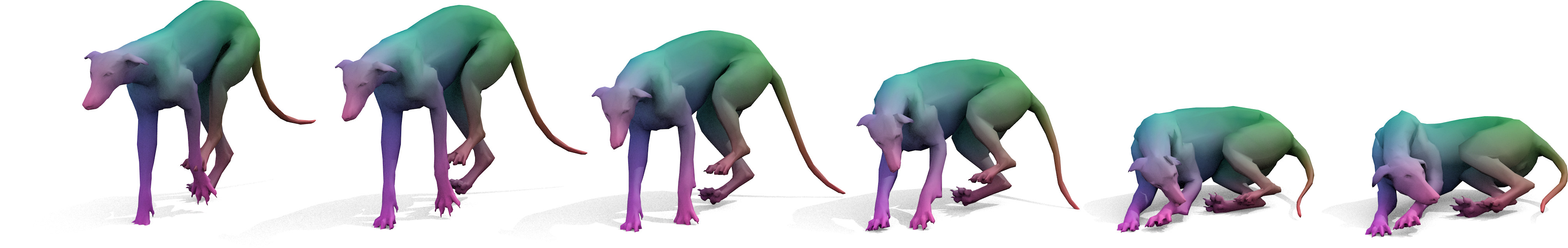}
\includegraphics[width=0.99\linewidth]{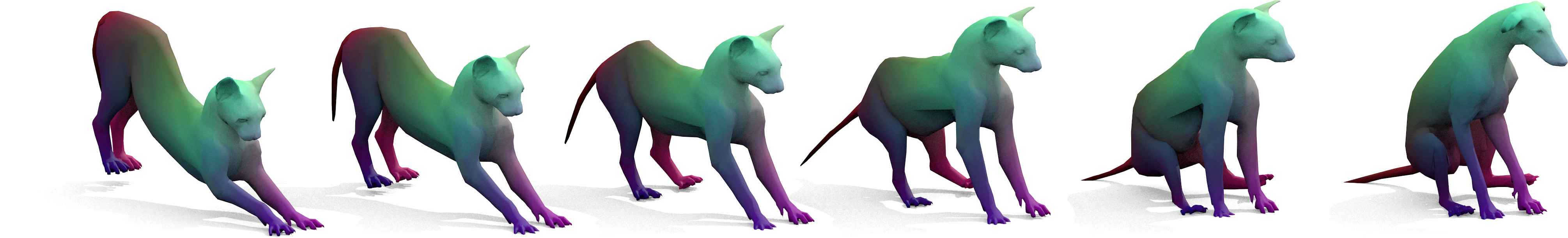}
\caption{\textbf{Interpolation on G-S-H.} Two interpolation sequences on our own benchmark G-S-H obtained with NeuroMorph. This shows clearly that, while our method contains interesting non-isometric pairs, the non-rigid pose variety is still significant.
}\label{fig:gshinterpol}
\end{figure*}

\begin{figure*}
\centering
\includegraphics[width=0.99\linewidth]{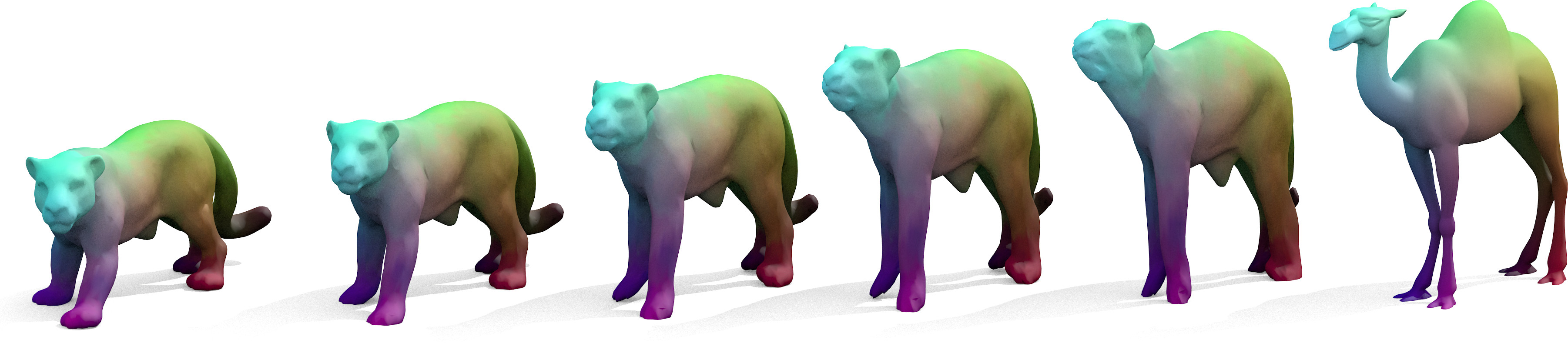}
\includegraphics[width=0.99\linewidth]{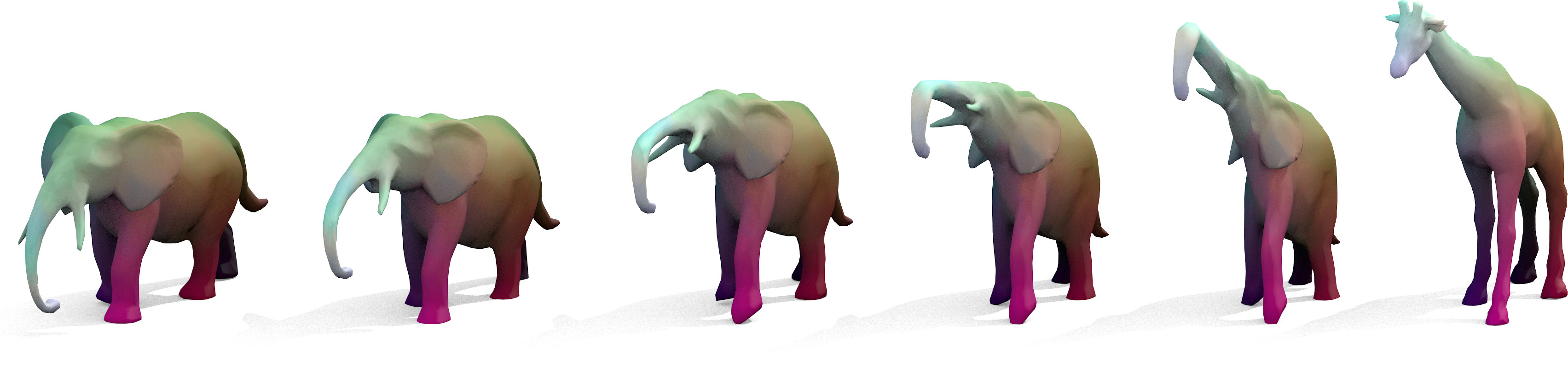}
\caption{\textbf{Interpolation on SHREC20.} We show two additional examples of interpolation sequences obtained with our method for pairs of shapes from the SHREC20~\cite{dyke2020shrec} dataset. For each input pair, our method acts on the pose of the first input objects (left) while mostly preserving its identity. The elephant uses its trunk to imitate the shape of the giraffe's head. While this can be considered meaningful from a geometric perspective, it also reveals a limitation of our approach. The fully unsupervised setup occasionally fails to find semantically exact correspondences, if the geometric features have a very different appearance.
}\label{fig:shrec20interpol}
\end{figure*}

\begin{figure*}
\centering
\begin{overpic}
        [width=0.82\linewidth]{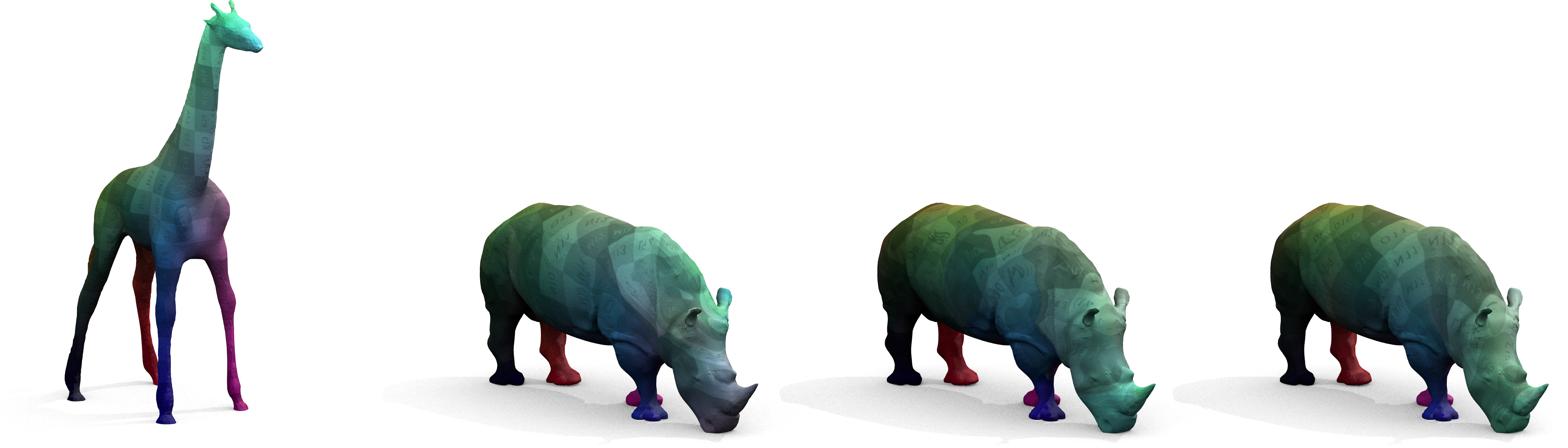}
        \put(10,32){Source\small}
        \put(30,32){Smooth Shells~\cite{eisenberger2020smooth}\small}
        \put(61,32){Ours\small}
        \put(83,32){Ours + SL\small}
\end{overpic}
\includegraphics[width=0.82\linewidth]{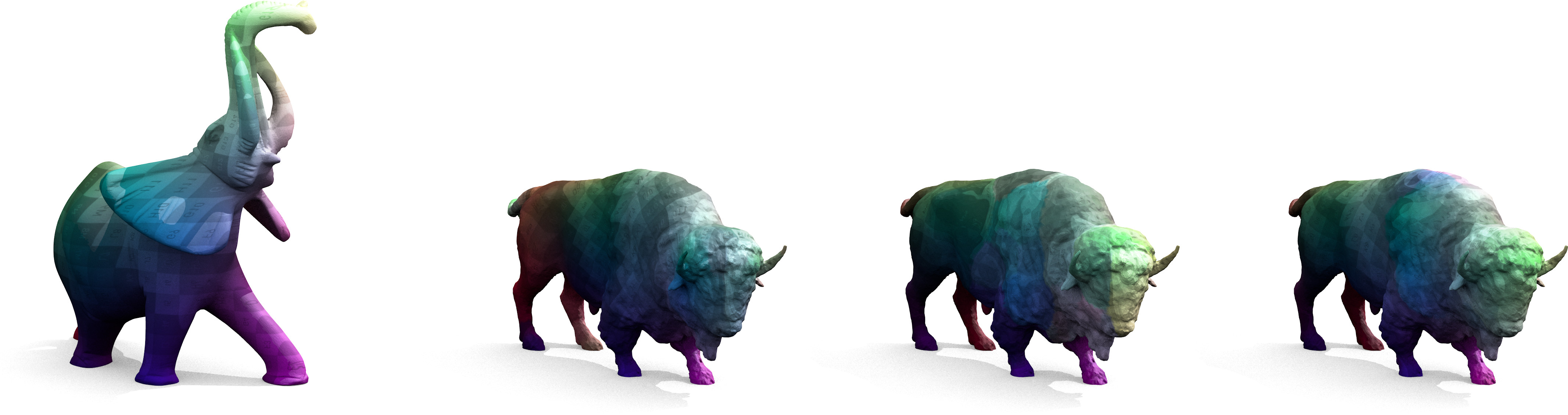}
\includegraphics[width=0.82\linewidth]{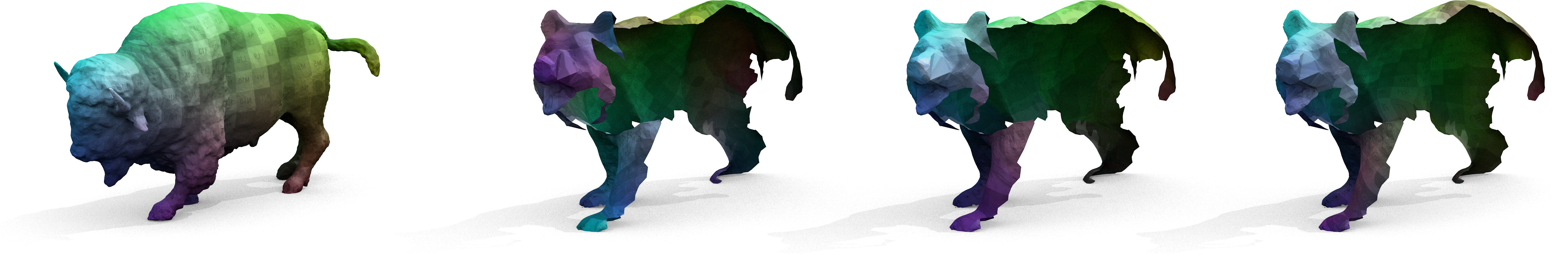}
\caption{\textbf{Unsupervised correspondences on SHREC20.} We show two more qualitative comparisons of correspondences obtained with different methods on the SHREC20 benchmark.
}\label{fig:shrec20qualappendix}
\end{figure*}

\begin{figure*}
\centering

\begin{overpic}
	[width=0.24\linewidth]{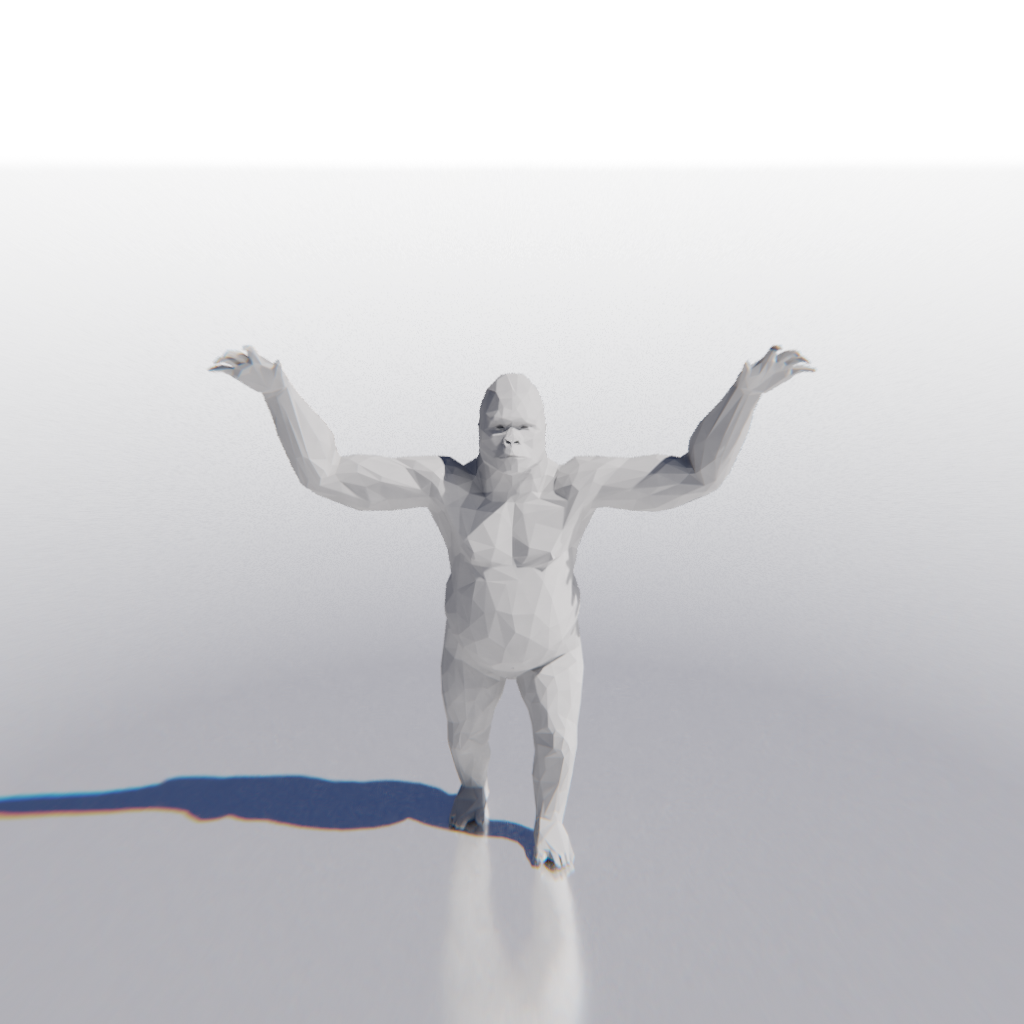}
       \put(40,86){Input $\mathcal{X}$\small}
\end{overpic}
\includegraphics[width=0.24\linewidth]{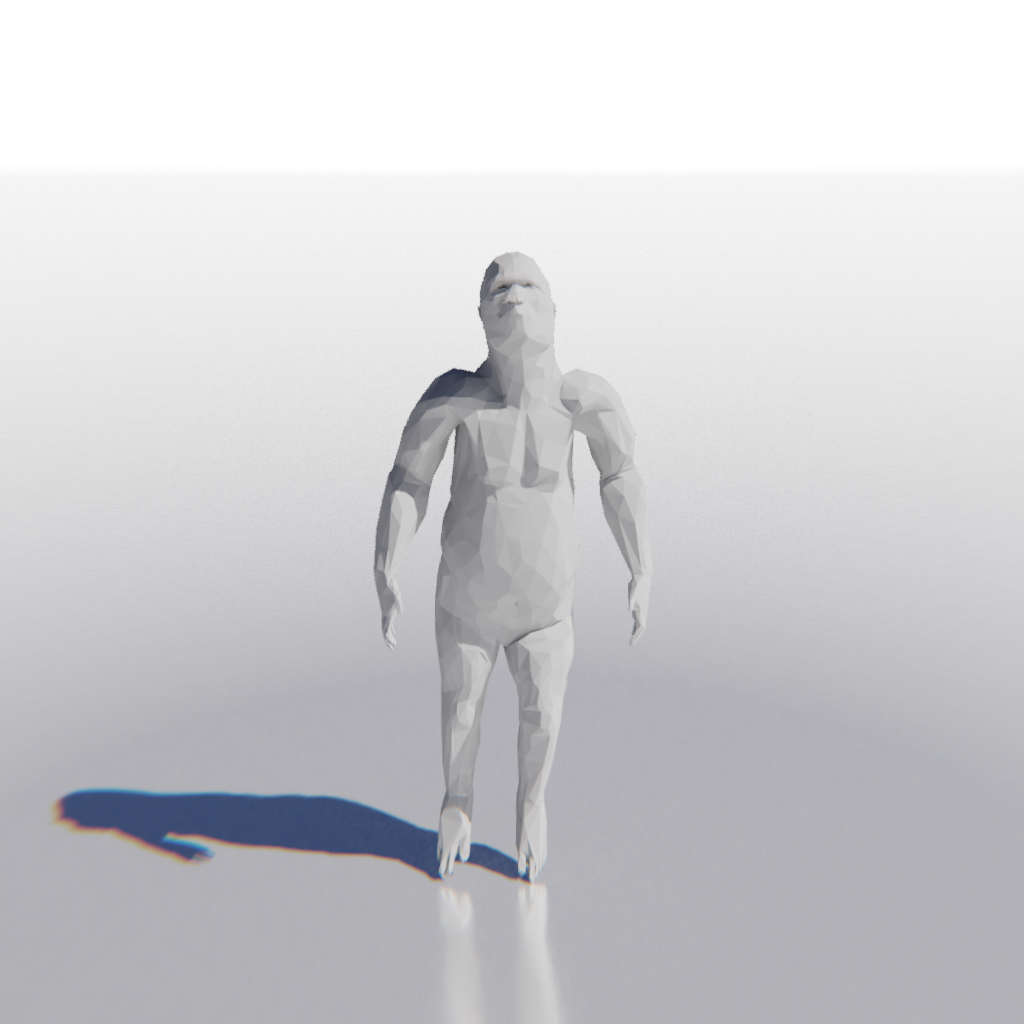}
\includegraphics[width=0.24\linewidth]{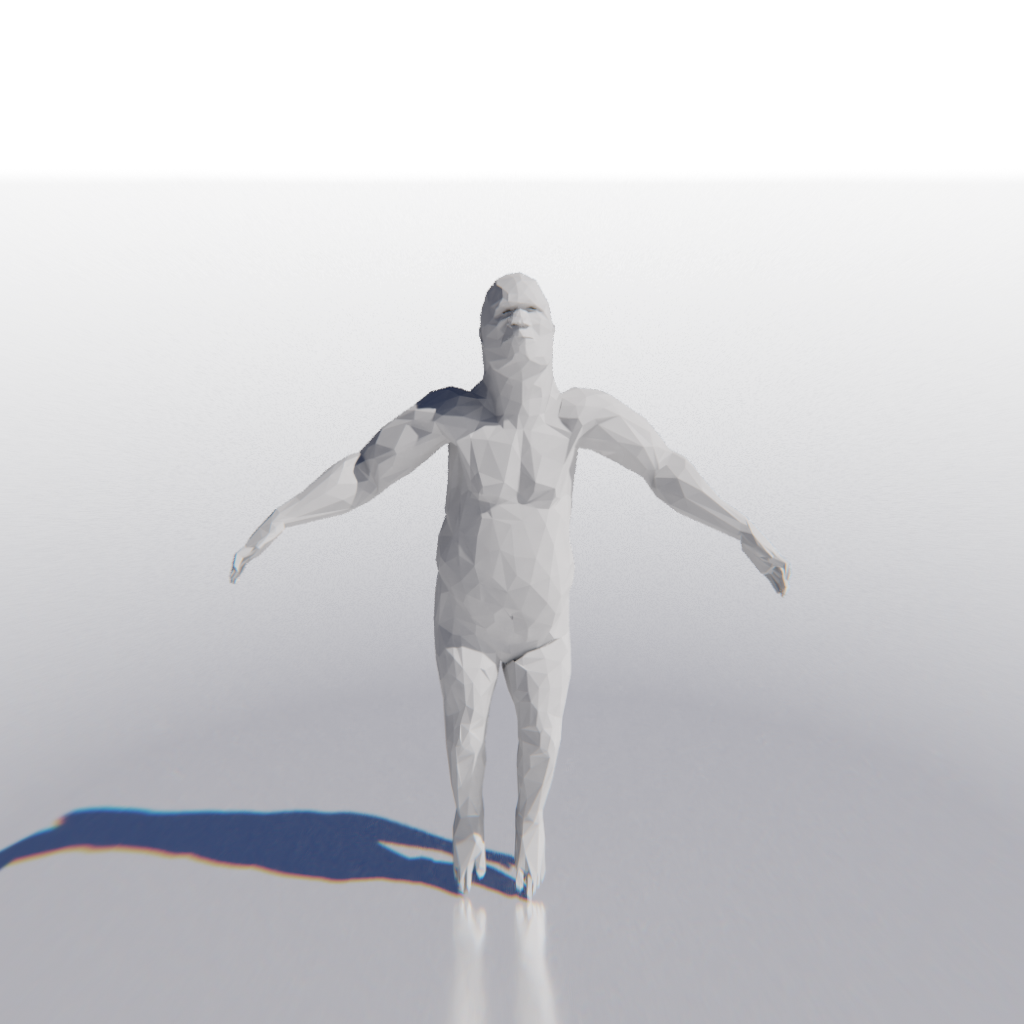}
\includegraphics[width=0.24\linewidth]{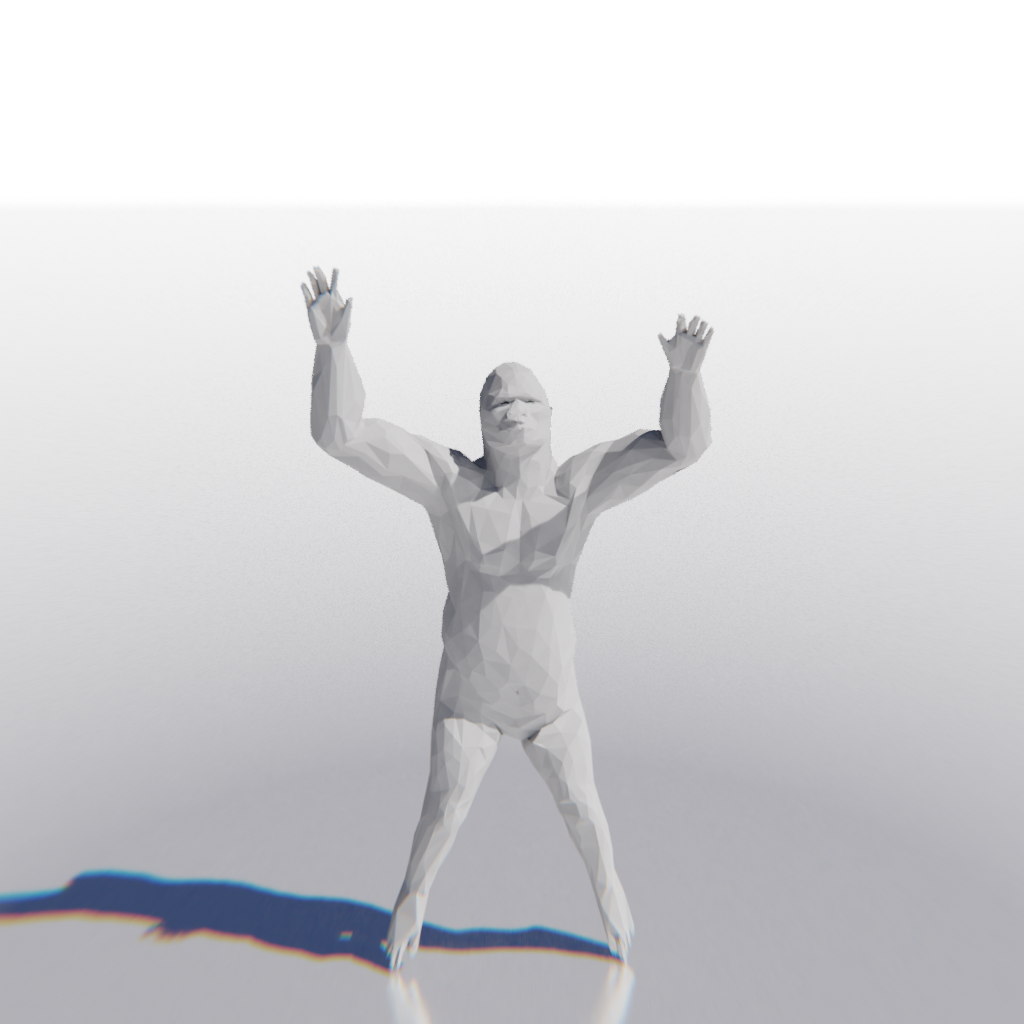}

\hspace{0.24\linewidth}
\includegraphics[width=0.24\linewidth]{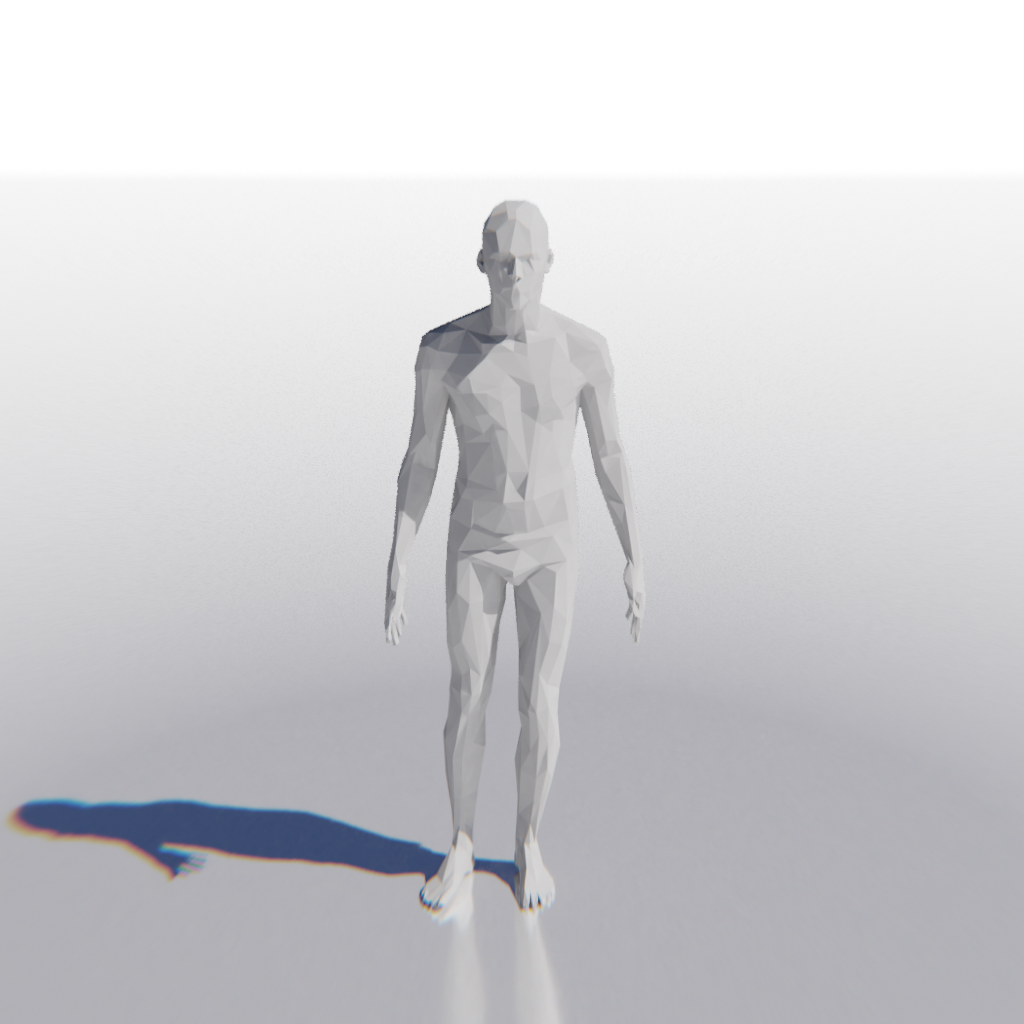}
\includegraphics[width=0.24\linewidth]{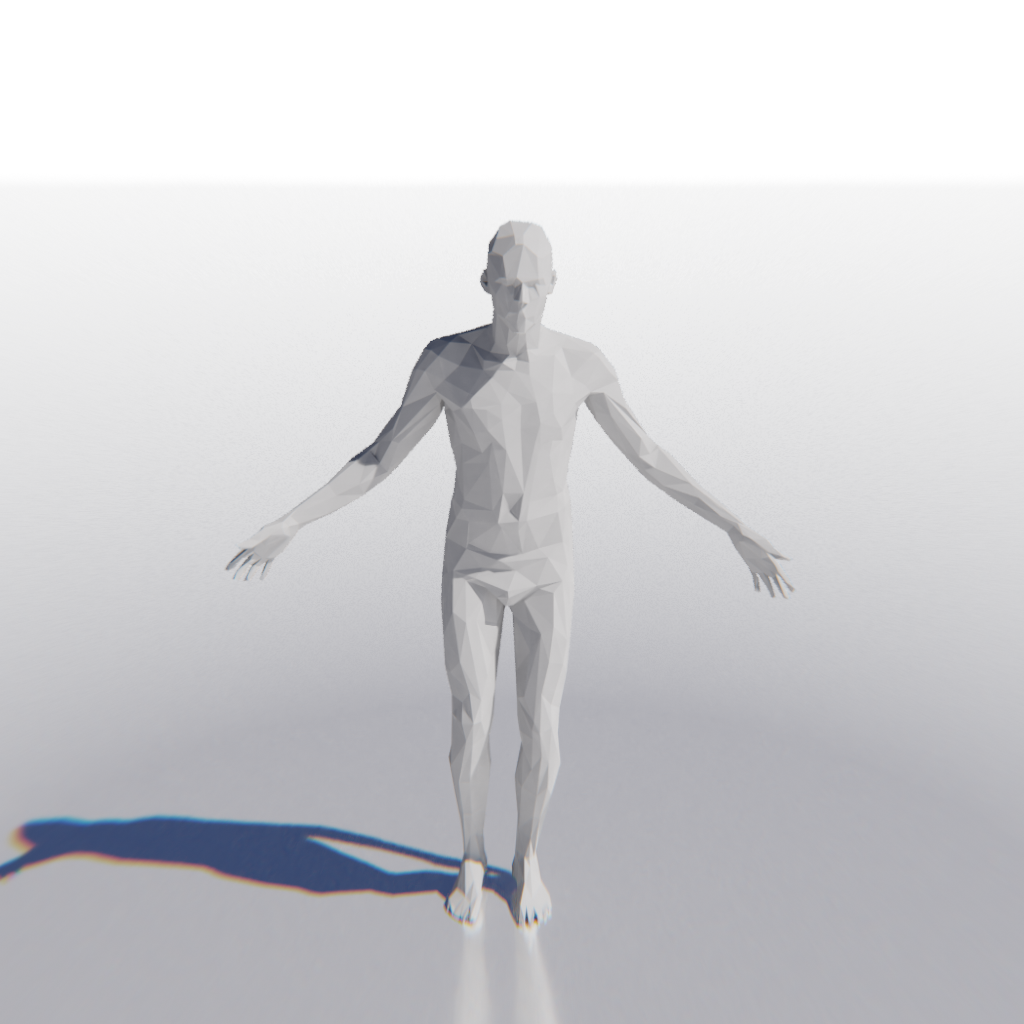}
\includegraphics[width=0.24\linewidth]{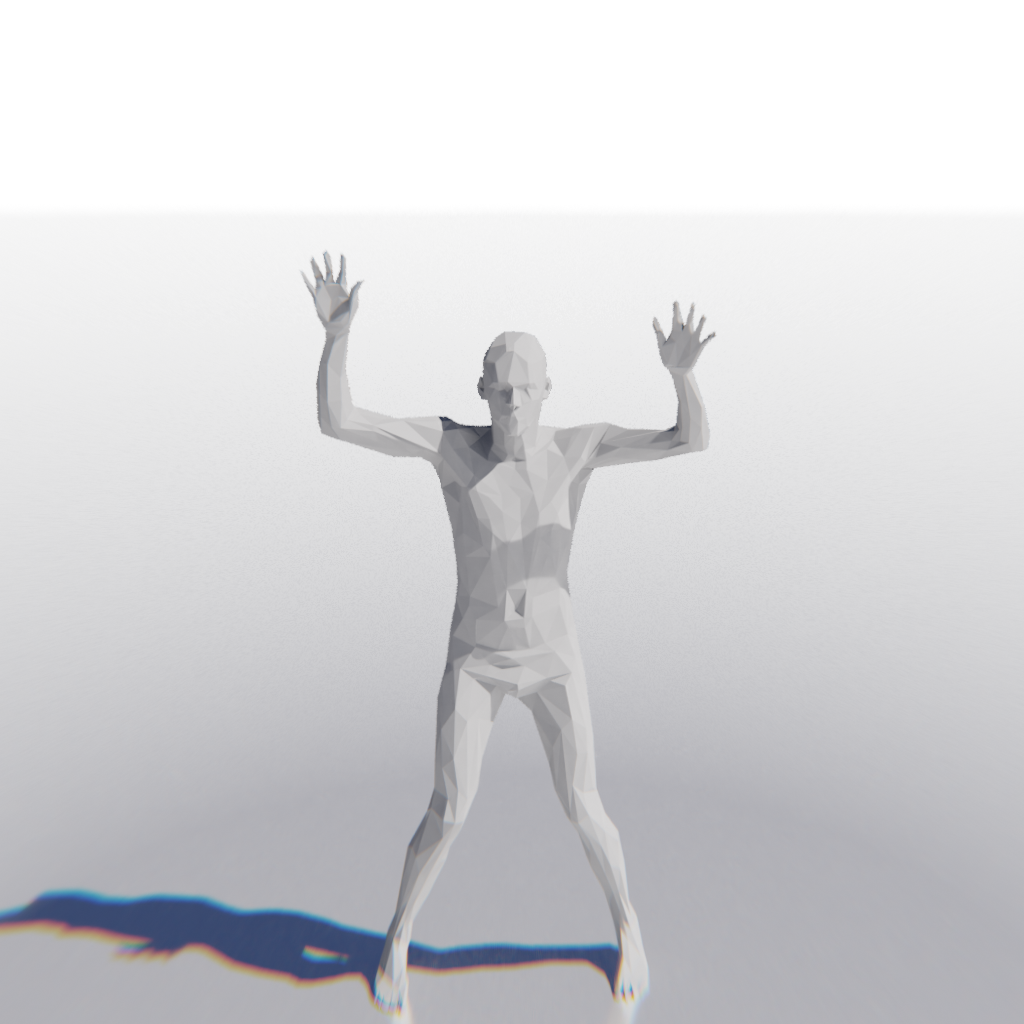}

\begin{overpic}
	[width=0.24\linewidth]{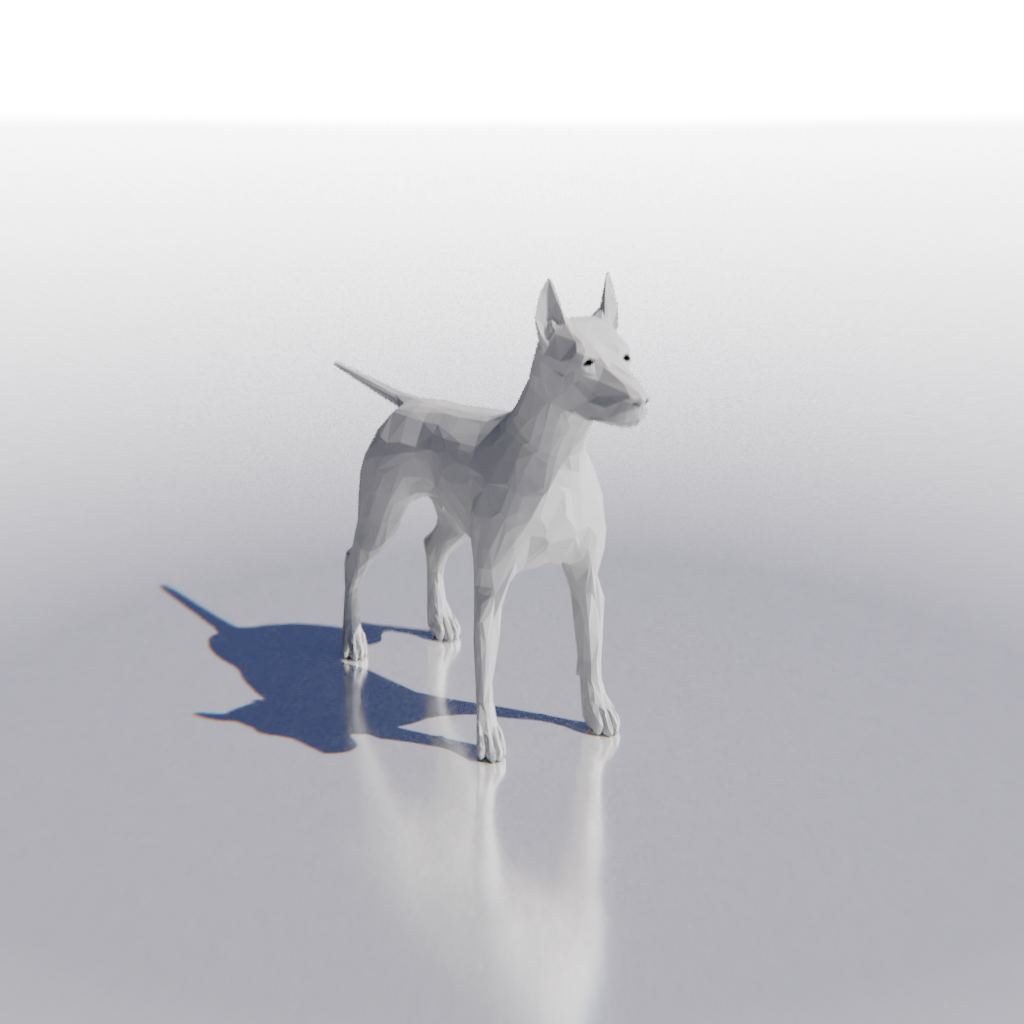}
       \put(40,89){Input $\mathcal{X}$\small}
\end{overpic}
\includegraphics[width=0.24\linewidth]{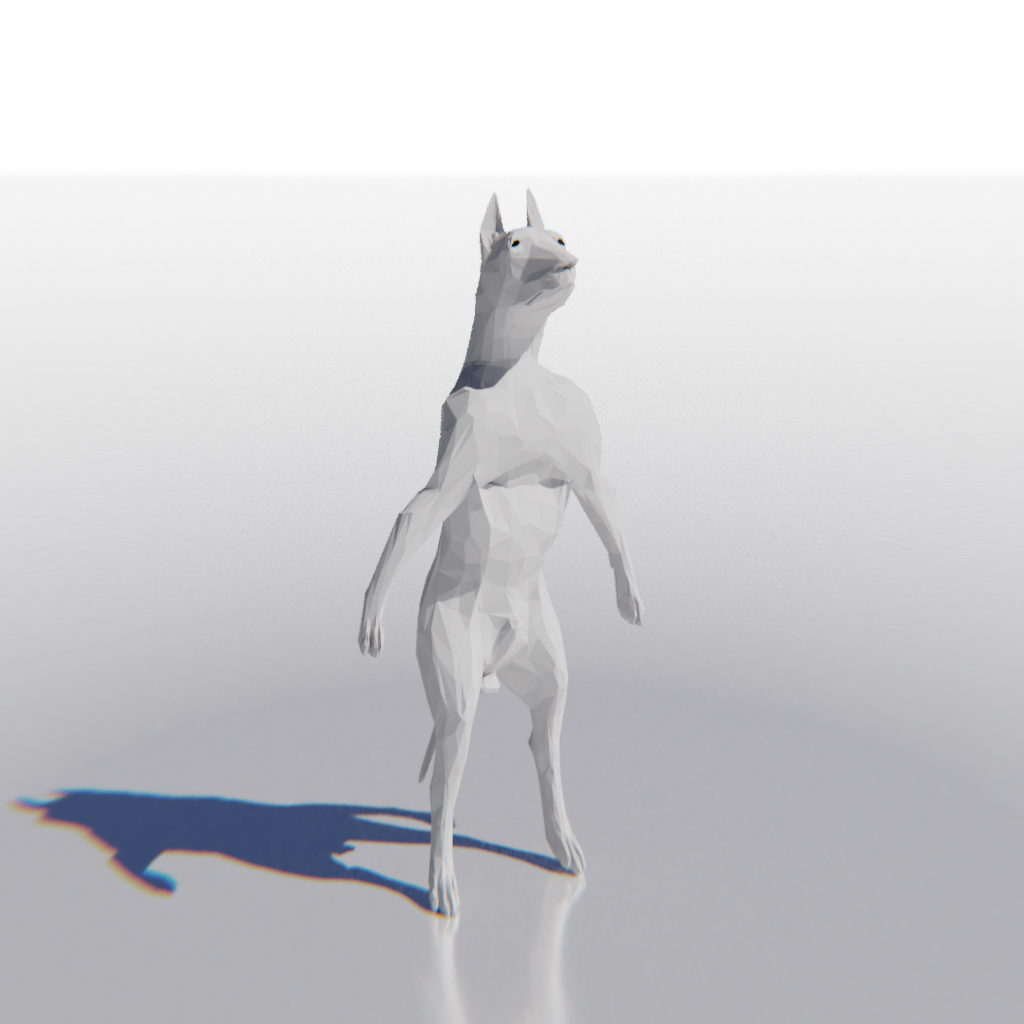}
\includegraphics[width=0.24\linewidth]{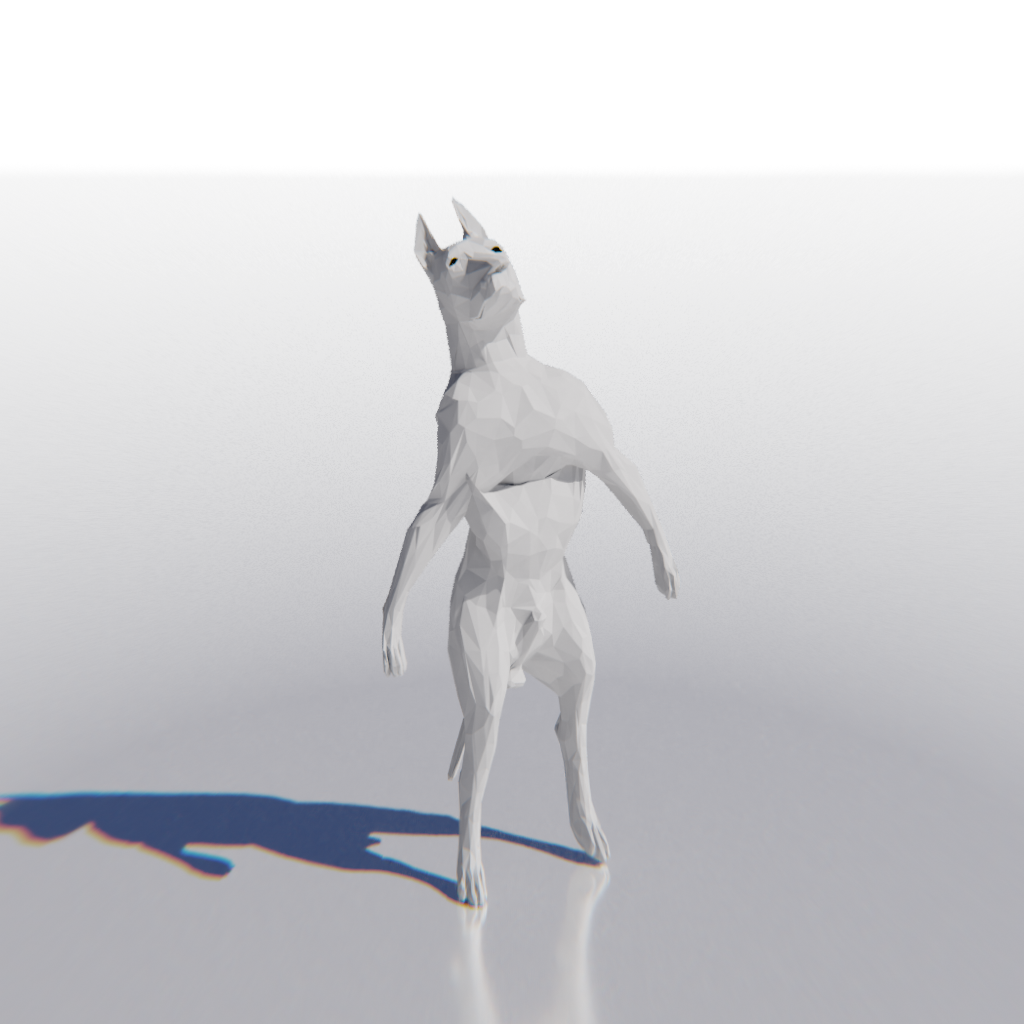}
\includegraphics[width=0.24\linewidth]{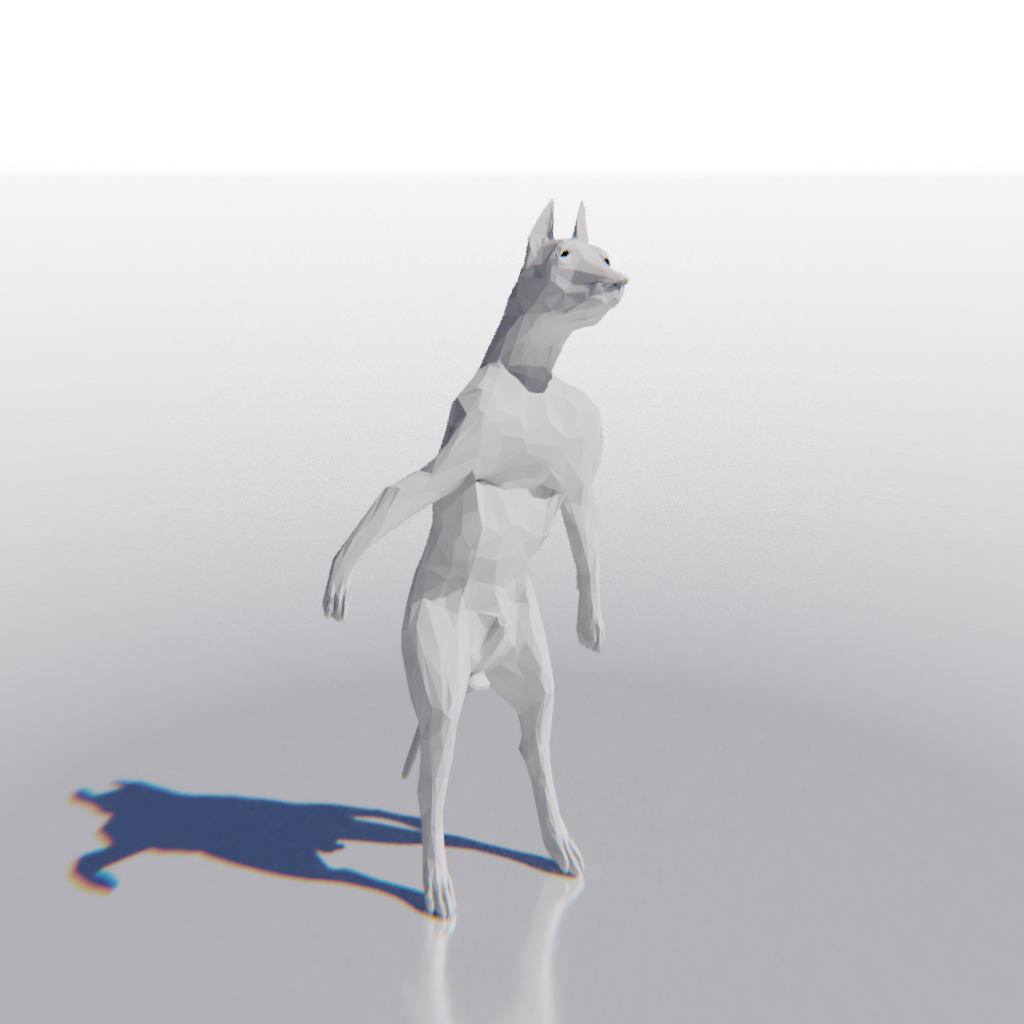}

\hspace{0.24\linewidth}
\includegraphics[width=0.24\linewidth]{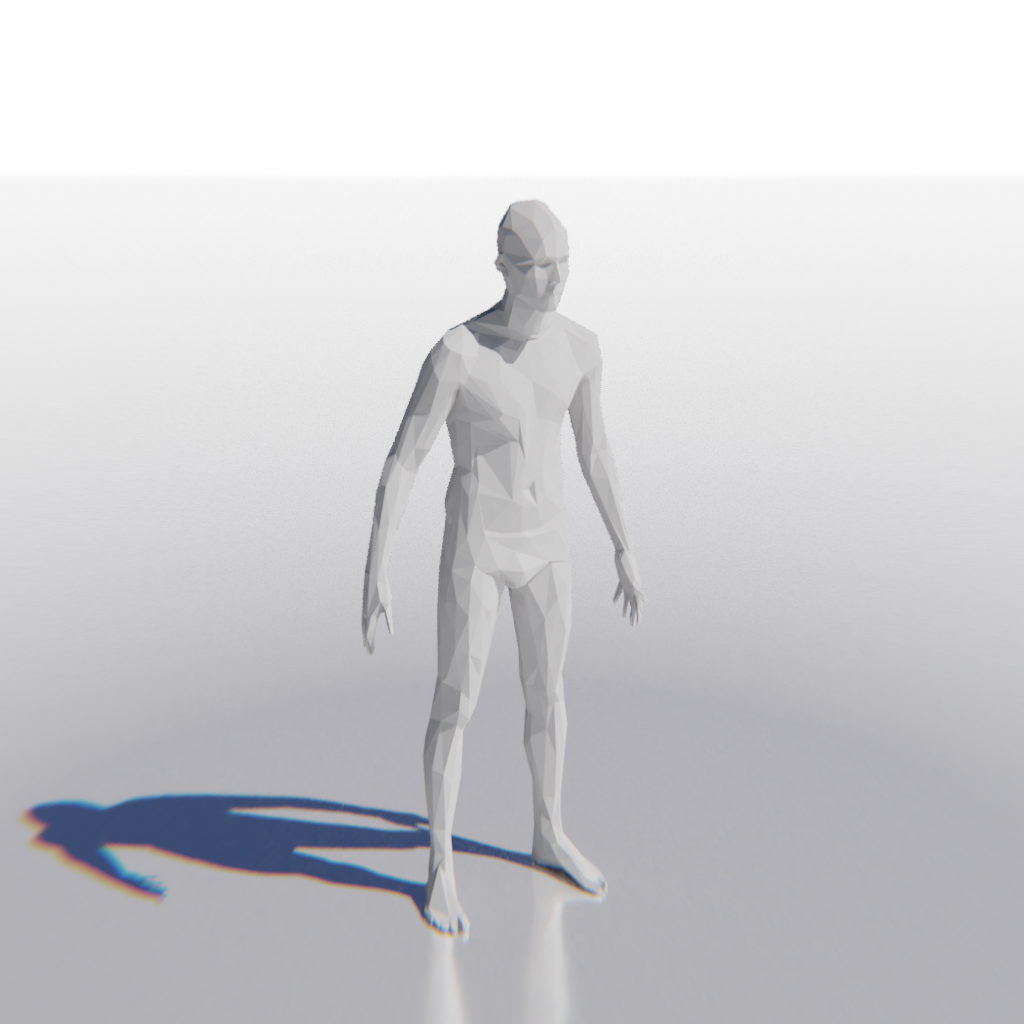}
\includegraphics[width=0.24\linewidth]{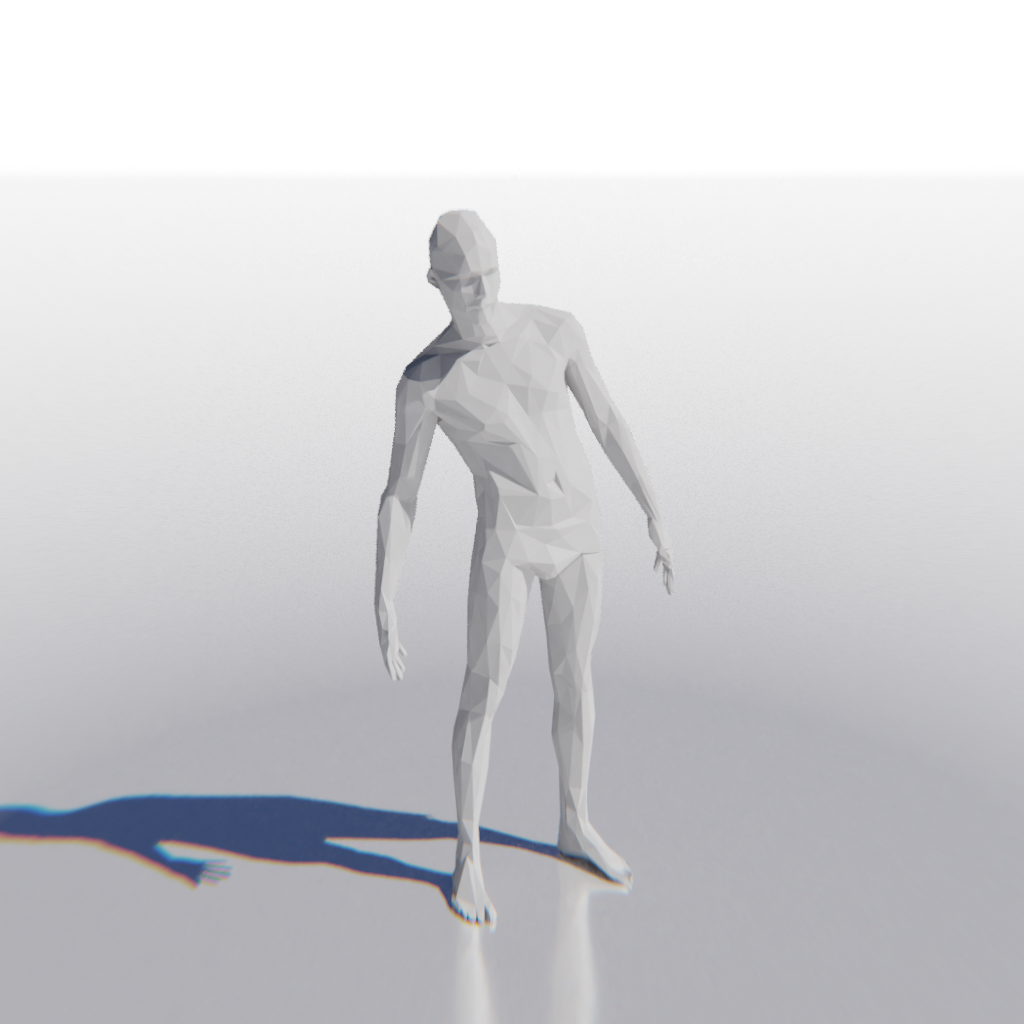}
\includegraphics[width=0.24\linewidth]{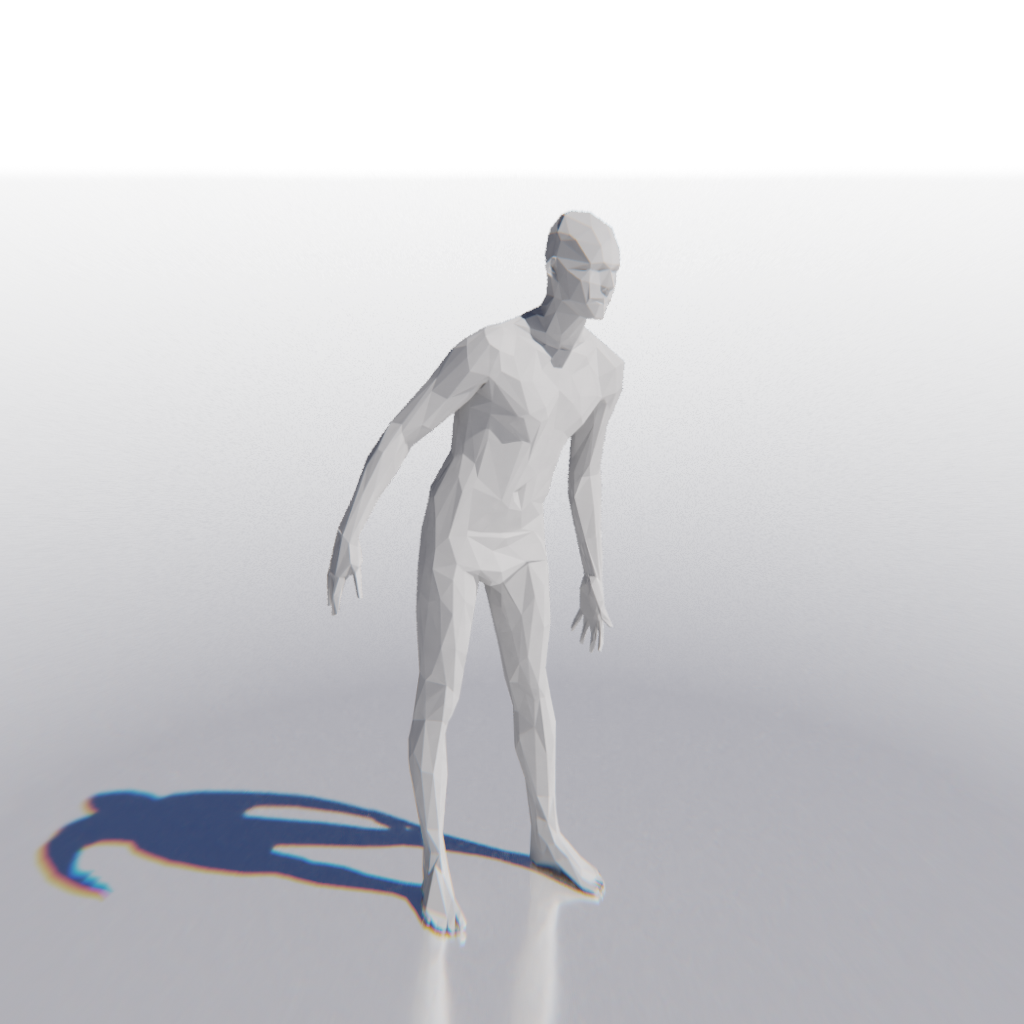}

\caption{\textbf{Digital puppeteering.} We train NeuroMorph jointly for a collection of animal and human (SMPL) shapes. In that manner, we effectively learn a pose prior for the animal shapes which allows us to animate them according to a reference sequence of SMPL shapes from DFAUST. See also our attached videos for the full, animated versions of the two sequences shown here.
}\label{fig:puppeteering}
\end{figure*}

In our experiments, we showed quantitative evaluations on multiple benchmarks, including FAUST~\cite{Bogo:CVPR:2014}, FAUST remeshed~\cite{ren2018orientation}, MANO~\cite{MANO:SIGGRAPHASIA:2017}, SURREAL~\cite{varol2017learning} and the SHREC20 challenge~\cite{dyke2020shrec}. These five, as well as many more existing 3D shape datasets, can be roughly classified in two classes: (i) Synthetic datasets with dense ground-truth, near-isometries or a compatible meshing and (ii) real datasets with non-isometric pairs and sparse annotated ground-truth correspondences\footnote{Of course not all existing benchmarks fall under one of these two categories. Some notable exceptions are datasets that specify on a certain class of objects (like humans)~\cite{melzi2019shrec,Bogo:CVPR:2014} or a specific type of input noise (like partiality or topological changes) \cite{rodola2016partial,laehner2016shrec}}. In many cases, for (i) the objects are within the same class and therefore have a similar intrinsic geometry, but they undergo challenging, extrinsic deformations with large, non-rigid pose discrepancies. For (ii), the topological proportions of a pair of objects can be quite different, but the poses are less challenging than (i).

To address this disconnect between non-isometries and large non-rigid deformations, in existing benchmarks we create our own dataset, where the goal is to jointly address all the challenges mentioned above: Our benchmark has non-isometric pairs of objects from different classes, large-scale non-rigid poses and dense annotated ground-truth correspondences for evaluation.

\subsection{The dataset}

We created objects of 3 different classes for our dataset with the tool ZBrush: A dog (Galgo), a cat (Sphynx) and a human. In modeling these shapes, we took great care to obtain generic but anatomically correct instances of these distinct species, see~\Cref{fig:gsh3x3} for example shapes from all three classes. We furthermore endowed all objects with a UV-map parameterization, as well as a wireframe acting as a deformation cage. Moreover, the range of motions of one object is specified by a hierarchical set of joints that is consistent for all objects in the dataset. We then animate the different objects by specifying different configurations in terms of deformation handles and applying the deformation to the full shapes with a skinning technique. The UV parameterizations were defined in a way that they are consistent across all considered classes, as a patchwork of smaller components/regions of all objects.

\subsection{Experiments}

We performed a number of experiments on our new benchmark. For evaluation, we select a number of $\sim 120$ uniformly sampled keyframes for training and define 32 different poses as our test set.  In the main paper, the matching accuracy for our method, as well as other unsupervised matching approaches are compared for this setup. Specifically, we followed the same evaluation protocol that we mentioned earlier in~\Cref{subsec:shapecorr} for the results in~\Cref{fig:heatmaps}. Since we have dense ground-truth correspondences that are consistent across all surfaces, we can also display the mean geodesic error at each individual point of the objects. Specifically, we take the UV-map parameterization on one pose of the `Galgo' shape from our dataset and display the mean matching error of all pairs in the test set.
Furthermore, we show qualitative examples of interpolations obtained with our method in~\Cref{fig:gshinterpol}.

\section{Ablation study}

We now provide an ablation study where we examine how certain parts of our method contribute to our empirical results. Specifically, we perform the following ablations:

\begin{enumerate}
\item[(i)] Remove the auxiliary correspondence loss $\ell_\mathrm{geo}$.
\item[(ii)] Train for correspondences directly without the interpolator module from our architecture (see \Cref{fig:eclayer}). This means that we only use $\ell_\mathrm{geo}$ and ignore the other two loss components.
\item[(iii)] Remove the max-pooling layers in~\Cref{eq:appendglobal} from our architecture.
\item[(iv)] Replace the EdgeConv layer in~\Cref{eq:edgeconv} with a standard PointNet~\cite{qi2017pointnet} layer.
\item[(v)] Replace our feature extractor with KPConv\footnote{KPConv is a state-of-the-art architecture for point cloud learning, but its main emphasis is on tasks like object classification and segmentation. It was, however, used in a matching pipeline before in prior work~\cite{donati2020deep}.}~\cite{thomas2019kpconv}.
\end{enumerate}

We then report how these changes affect the geodesic error and the mean conformal distortion (interpolation error) on FAUST remeshed, corresponding to the results in~\Cref{table:matchingaccuracy} and~\Cref{fig:manofaustcomparison}. Specifically, we compare the results without post-processing:


\begin{table}[!h]
\begin{tabular}{clcc}
\toprule
{}    &                      & Geo. err.    & Conf. dist\\
\midrule
Ours  &                      & \textbf{2.3} & \textbf{0.10} \\
(i)   & No $\ell_\text{geo}$ & 13.0         & 0.13\\
(ii)  & No interp.           & 4.7          & -- \\
(iii) & No maxpool           & 2.5          & 0.14\\
(iv)  & EdgeConv             & 10.6         & 0.25 \\
(v)   & Use KPConv           & 4.2          & 0.28 \\
\bottomrule
\end{tabular}
\centering
\caption{Ablations.}\label{table:ablation}
\end{table}

These experiments indicate that both the interpolator and the feature extractor are crucial for obtaining high quality results: Modifying technical details of our feature extractor leads to suboptimal results (iii)-(v). 
The difference is particularly large when EdgeConv is replaced by PointNet (iv).
Similarly, without the interpolator module, the correspondence estimation is less accurate (ii), since they are not based on an explicit notion of extrinsic deformation. Finally, without the geodesic loss $\ell_\mathrm{geo}$, the matching accuracy deteriorates significantly (i). This can be attributed to the fact that, without a notion of intrinsic geometry, our method is prone to run into unmeaningful local minima.

\section{Additional qualitative examples}

Finally, we show a few more qualitative results from the SHREC20 benchmark. Specifically, we display examples of non-isometric interpolations in~\Cref{fig:shrec20interpol} and a qualitative comparison of correspondences obtained with our method and smooth shells~\cite{eisenberger2020smooth} in~\Cref{fig:shrec20qualappendix}.

\section{Digital puppeteering}

One interesting property of our method is that it is able to learn geometrically plausible pose priors for any shape $\mathcal{X}$. Given any target pose $\mathcal{Y}$, we generally obtain a meaningful new pose of the input object $\mathcal{X}$ as the last pose of the interpolation sequence $t=1$. Consequently, by considering a distribution of target poses $\mathcal{Y}$, we automatically obtain a shape space of admissible poses with the object identity $\mathcal{X}$. This allows for digital puppeteering as an application of our method. To that end, we jointly train NeuroMorph for a set of poses from the TOSCA dataset of animals and humans, as well as the SURREAL dataset which consists of a large collection of SMPL shapes. As a proof of concept, we then query our network for a time-continuous sequence of SMPL shapes from the DFAUST dataset and animate the sequence by replacing the human shape with different animals, see~\Cref{fig:puppeteering} and also see our attached videos in the supplementary material. 

\section{Additional quantitative comparisons}

\begin{figure}
\includegraphics[width=\linewidth]{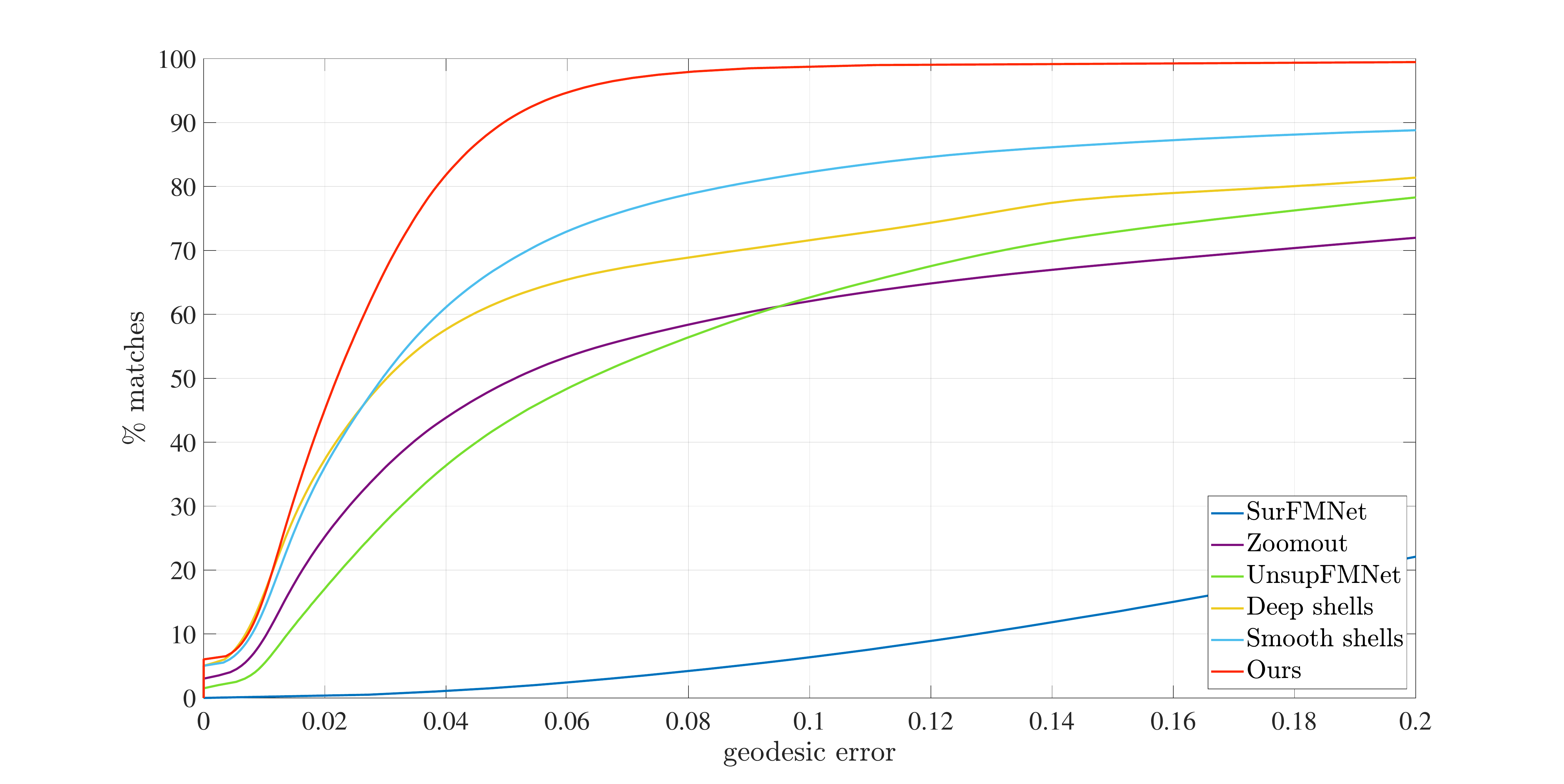}
\caption{\textbf{Unsupervised correspondences on SHREC19 remeshed.}
A comparison of unsupervised approaches, showing the cumulative geodesic error curves on the 430 challenge test pairs.}\label{fig:shrec19comparison}
\end{figure}

For the sake of completeness, we also provide quantitative comparisons on the SHREC19~\cite{melzi2019shrec} benchmark, see \Cref{fig:shrec19comparison}. Note that, like for FAUST, we again use the more recent remeshed version of the dataset, first introduced in~\cite{donati2020deep}.

\end{document}